\documentclass[10pt,twocolumn,letterpaper]{article}

\usepackage[pagenumbers]{cvpr} 
\usepackage[accsupp]{axessibility}  
\usepackage{graphicx}
\usepackage{cite}
\usepackage{amsmath}
\usepackage{amssymb}
\usepackage{booktabs}
\usepackage{times}
\usepackage{epsfig}
\DeclareMathOperator*{\argmin}{arg\,min}
\DeclareSymbolFont{matha}{OML}{txmi}{m}{it}
\DeclareMathSymbol{\varv}{\mathord}{matha}{118}
\usepackage{array}
\usepackage{tabularx}
\usepackage{multirow, booktabs}
\usepackage{float}
\usepackage{bm}
\usepackage[dvipsnames,table,xcdraw]{xcolor}
\usepackage{resizegather}

\usepackage[pagebackref,breaklinks,colorlinks]{hyperref}

\def\F{\mathbf{F}}
\def\M{\mathbf{M}}
\def\h{\bm{h}}

\usepackage[capitalize]{cleveref}
\crefname{section}{Sec.}{Secs.}
\Crefname{section}{Section}{Sections}
\Crefname{table}{Table}{Tables}
\crefname{table}{Tab.}{Tabs.}

\usepackage[dvipsnames]{xcolor}

\definecolor{cvprblue}{rgb}{0.21,0.49,0.74}

\def\paperID{2412} 
\def\confName{CVPR}
\def\confYear{2024}

\title{Modality-Agnostic Structural Image Representation Learning for \\ Deformable Multi-Modality Medical Image Registration}

\author{Tony C. W. Mok\textsuperscript{1,2}\footnotemark[1] \quad Zi Li\textsuperscript{1,2}\footnotemark[1] \quad Yunhao Bai\textsuperscript{1} \quad Jianpeng Zhang\textsuperscript{1,2,4} \quad Wei Liu\textsuperscript{1,2} \\ \quad Yan-Jie Zhou\textsuperscript{1,2,4} \quad Ke Yan\textsuperscript{1,2} \quad Dakai Jin\textsuperscript{1} \quad Yu Shi\textsuperscript{3} \quad Xiaoli Yin\textsuperscript{3} \quad Le Lu\textsuperscript{1} \quad Ling Zhang\textsuperscript{1}\\
\normalsize{\textsuperscript{1} DAMO Academy, Alibaba Group} \\
\normalsize{\textsuperscript{2} Hupan Lab, 310023, Hangzhou, China} \\
\normalsize{\textsuperscript{3} Shengjing Hospital of China Medical University, China}\\
\normalsize{\textsuperscript{4} College of Computer Science and Technology, Zhejiang University, China}\\
{\tt\small cwmokab@connect.ust.hk}\\
}

\begin{document}
\maketitle

\renewcommand*{\thefootnote}{\fnsymbol{footnote}}
\setcounter{footnote}{1}
\footnotetext{Contributed equally.}
 
\begin{abstract}

Establishing dense anatomical correspondence across distinct imaging modalities is a foundational yet challenging procedure for numerous medical image analysis studies and image-guided radiotherapy. Existing multi-modality image registration algorithms rely on statistical-based similarity measures or local structural image representations. However, the former is sensitive to locally varying noise, while the latter is not discriminative enough to cope with complex anatomical structures in multimodal scans, causing ambiguity in determining the anatomical correspondence across scans with different modalities. In this paper, we propose a modality-agnostic structural representation learning method, which leverages Deep Neighbourhood Self-similarity (DNS) and anatomy-aware contrastive learning to learn discriminative and contrast-invariance deep structural image representations (DSIR) without the need for anatomical delineations or pre-aligned training images. We evaluate our method on multiphase CT, abdomen MR-CT, and brain MR T1w-T2w registration. Comprehensive results demonstrate that our method is superior to the conventional local structural representation and statistical-based similarity measures in terms of discriminability and accuracy.

\end{abstract}    
\section{Introduction}\label{sec:intro}

\begin{figure}[t]
	\begin{center}
        \includegraphics[width=1.0\linewidth]{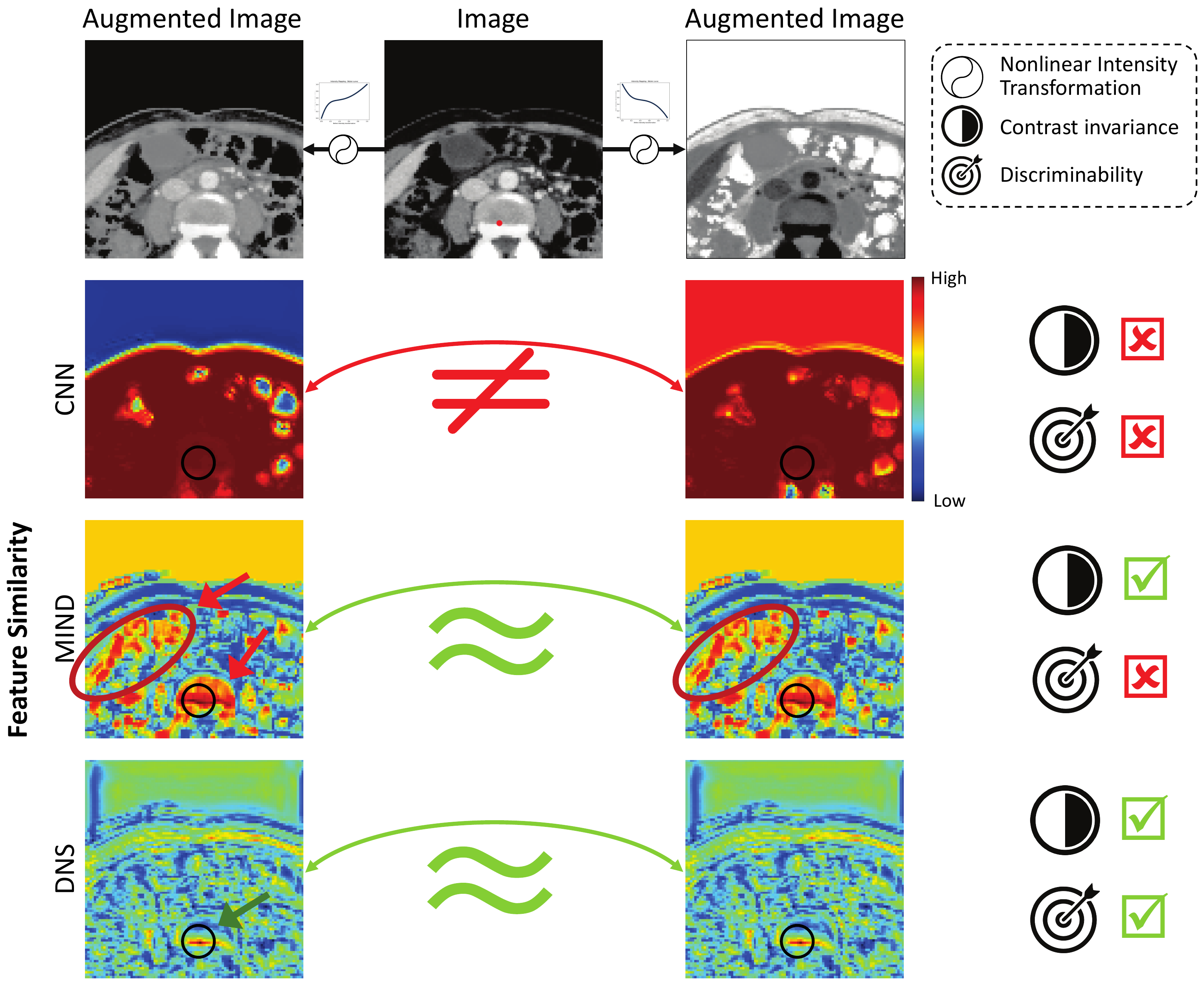}
	\end{center}

 \vspace{-0.8em} 
    \caption{Visualization of feature similarity between the marked feature vector (red dot) of the image and all feature vectors of augmented images using the convolutional neural network without pertaining (CNN), Modality Independent Neighbourhood Descriptor (MIND), and our proposed Deep Neighbourhood Self-similarity (DNS). Our method captures the contrast invariant and high discriminability structural representation of the image, reducing the ambiguity in matching the anatomical correspondence between multimodal images.}
     \vspace{-1.em}

	\label{fig: compare}
\end{figure}

Determining anatomical correspondence between multimodal data is crucial for medical image analysis and clinical applications, including diagnostic settings \cite{maurer1993review}, surgical planning \cite{alam2018medical,xiao2019evaluation} and post-operative evaluation \cite{mok2022unsupervised}. As a vital component for modern medical image analysis studies and image-guided interventions, deformable multimodal registration aims to establish the dense anatomical correspondence between multimodal scans and fuse information from multimodal scans, e.g., propagating anatomical or tumour delineation for image-guided radiotherapy \cite{lu2011integrated}. Since different imaging modalities provide valuable complementary visual cues and diagnosis information of the patient, precise anatomical alignment between multimodal scans benefits the radiological observation and the subsequent downstream computerized analyses. However, finding anatomical correspondences between homologous points in multimodal images is notoriously challenging due to the complex appearance changes across modalities. For instance, in multiphase abdomen computed tomography (CT) scans, the soft tissues can be deformed due to gravity, body motion, and other muscle contractions, resulting in an unavoidable large non-linear misalignment between subsequent imaging scans. Moreover, anatomical structures and tumours in multiphase CT scans show heterogeneous intensity distribution across different multiphase contrast-enhanced CT scans due to the intravenously injected contrast agent during the multiphase CT imaging.

Despite there being vast research studies \cite{mok2022unsupervised,hoopes2021hypermorph,mok2020fast,falta2022learning,shu2021medical,LiuLFZHL22,LiuLZFL20,bigalke2023unsupervised} on deformable image registration, most of these are focused on the mono-modal registration settings and rely on intensity-based similarity metrics, \ie, normalized cross-correlation (NCC) and mean squared error (MSE), which are not applicable to the multimodal registration.  Recently, several methods have proposed to learn an inter-domain similarity metric using supervised learning with pre-aligned training images \cite{gutierrez2017guiding,lee2009learning,pielawski2020comir}. However, the perfectly aligned images and the ideal ground truth deformations are often absent in multimodal medical images, which limits the applicability of these methods.

Historically, a pioneering work of Maes~\emph{et~al.}~\cite{maes1997multimodality} uses mutual information (MI) \cite{viola1997alignment} to perform rigid multimodal registration. Nevertheless, for deformable multimodal registration, many disadvantages have been identified when using the MI-based similarity measures \cite{pluim2000image}. Specifically, MI-based similarity measures are sensitive to locally varying noise distribution but not sensitive to the subtle anatomical and vascular structures due to the statistical nature of MI. 

As an alternative to directly assessing similarity or MI on the original images, structural image representation approaches have gained great interest for deformable multimodal registration. By computing the intermediate structural image representation independent of the underlying image acquisition, well-established monomodal optimization techniques can be employed to address the multimodal registration problem. A prominent example is the Modality-Independent Neighbourhood Descriptor~\cite{HeinrichJBMGBS12}, which is motivated by image self-similarity \cite{shechtman2007matching} and able to capture the internal geometric layouts of local self-similarities within images. Yet, such local feature descriptors are not expressive and discriminative enough to cope with complex anatomical structures in abdomen CT, exhibiting many local optima, as shown in Fig. \ref{fig: compare}. Therefore, it is often jointly used with a dedicated optimization strategy or requires robustness initialization. 


In this paper, we analyze and expose the limitations of self-similarity-based feature descriptors and mutual information-based methods in multi-modality registration. 
We depart from the classical self-similarity descriptor and propose a novel structural image representation learning paradigm dedicated to learning expressive deep structural image representations (DSIRs) for deformable multimodal registration. Our proposed method reduces the multimodal registration problem to a monomodal one, in which existing well-established monomodal registration algorithms can be applied. To the best of our knowledge, this is the first modality-agnostic structural representation learning approach that learns to capture DSIR with high discriminability from multimodal images without using perfectly aligned image pairs or anatomical delineation.

The main contributions of this work are as follows:
\begin{itemize}
 \item we propose a novel self-supervised deep structural representation learning approach for multimodal image registration that learns to extract deep structural image representation from standalone medical images, circumventing the need for anatomical delineations or perfectly aligned training image pair for supervision.

 \item we introduce the Deep Neighbour Self-similarity (DNS), which can capture long-range and complex structural information from medical images, addressing the ambiguity in classical feature descriptors and similarity metrics.
 
 \item we propose a novel contrastive learning strategy with non-linear intensity transformation, maximizing the discriminability of the feature representation across anatomical positions with homogeneous and heterogeneous intensity distribution.

 \item we demonstrate that the proposed deep structural image representation can be adapted to a variety of well-established learning-based and iterative optimization registration algorithms, reducing the multimodal registration problem to a monomodal registration problem.
\end{itemize}

We rigorously evaluate the proposed method on three challenging multimodal registration tasks:  liver multiphase CT registration, abdomen magnetic resonance imaging (MR) to CT registration, and brain MR T1w-T2w registration. Results demonstrate that our method is capable of computing highly expressive and discriminative structural representations of multimodal images, reaching the state-of-the-art performance of conventional methods solely with a simple gradient decent-based optimization framework.

\section{Related Work}

\paragraph{Multi-modal image registration.}
In general, multi-modal registration methods can be divided into three categories: statistical-based, structural representation-based and deep learning-based methods.

Prominent examples of statistical-based methods use information theory and optimize joint voxel statistics, such as minimizing the MI or normalized mutual information (NMI) as similarity measures for multi-modal registration~\cite{avants2009advanced,AvantsTSCKG11,SunNK14}. These approaches aim to estimate a solution that minimizes the entropy of the joint histogram between image pairs for rigid registration. However, as evidenced in \cite{pluim2000image}, MI-based similarity measures are restricted to measuring the statistical co-occurrence of image intensities, which are not extendable to the co-occurrence of complex patterns, such as subtle structural information of soft tissues and vascular structures in medical images, and are sensitive to locally varying noise distribution, which is not ideal for deformable multi-modal registration.




To circumvent the limitation of statistical-based methods, another common strategy is to project multimodal images into common intermediate structural representations or measure the misalignment through the modality-invariant similarity metrics~\cite{LiuCHCRJXYLLH20,LiuLFZHL22,10231109}. Prominent examples include MIND ~\cite{HeinrichJBMGBS12,HeinrichJBS12,heinrich2013towards}, attribute vectors~\cite{shen2002hammer} and Linear Correlation of Linear Combination (LC$^2$) \cite{wein2008automatic} that are designed to capture a dense structural image representation of multi-modal images that are independent of the underlying image acquisition. However, these local structural representations are often not sufficiently discriminative and expressive to drive a non-rigid registration with many degrees of freedom, not differentiable or expensive to compute. As such,  these local structural representations are often used in conjunction with robust optimization methods, \eg discrete optimization \cite{heinrich2012globally} and Bound Optimization by Quadratic Approximation~\cite{powell2009bobyqa}, or requiring robust rigid initialization.



Deep learning-based image registration (DLIR) methods have demonstrated remarkable results on diverse mono-modal and multi-modal registration tasks, as evidenced by tremendous registration benchmarks \cite{hering2022learn2reg,eisenmann2022biomedical}. However, the success of recent DLIR approaches has largely been fueled by the supervision of anatomical segmentation labels \cite{muller2014deriving,zeng2022learning} or the supervision of perfectly aligned multimodal images \cite{gutierrez2017guiding,lee2009learning,pielawski2020comir}. The absence of pre-aligned multimodal medical images and dependence on segmentation labels further restricts their generalizability across different anatomies or modalities. 
In contrast to the mainstream DLIR and learning-based structural representation methods, our proposed method is fully self-supervised, which circumvents the need for anatomical delineations or perfectly aligned multi-modal images.


\paragraph{Contrastive learning in image registration.}
Motivated by the success of contrastive learning in visual representation learning \cite{henaff2020data,he2020momentum,chen2020simple}, several methods \cite{YanCJMGHTXLL22,LiTMBWGZLYYJ23,abs-2307-03535} adopt contrastive learning to extract anatomical structural embedding for monomodal registration. Contrastive learning focuses on extracting discriminative representations by contrasting positive and negative pairs of instances~\cite{He0WXG20}. These methods use noise contrastive estimation (NCE)~\cite{abs-1807-03748}, learning an anatomical structural representation where the feature vectors from the same anatomical location are brought together, in contrast to feature vectors from different anatomical locations. Nevertheless, the learned anatomical structural representations of these methods are not contrast invariance, hence, incapable of multimodal registration. Apart from learning structural representation, a recent work \cite{DeySSZGS22} directly minimizes the PatchNCE \cite{park2020contrastive} loss for brain MR T1w-T2w registration. Yet, minimizing the PatchNCE is identical to maximizing a lower bound on mutual information between corresponding spatial locations in the feature maps, which inherits the limitation of MI in deformable registration.

\section{Method}\label{sec:formatting}

\begin{figure*}[t]
	\begin{center}
        \includegraphics[width=.8\linewidth]{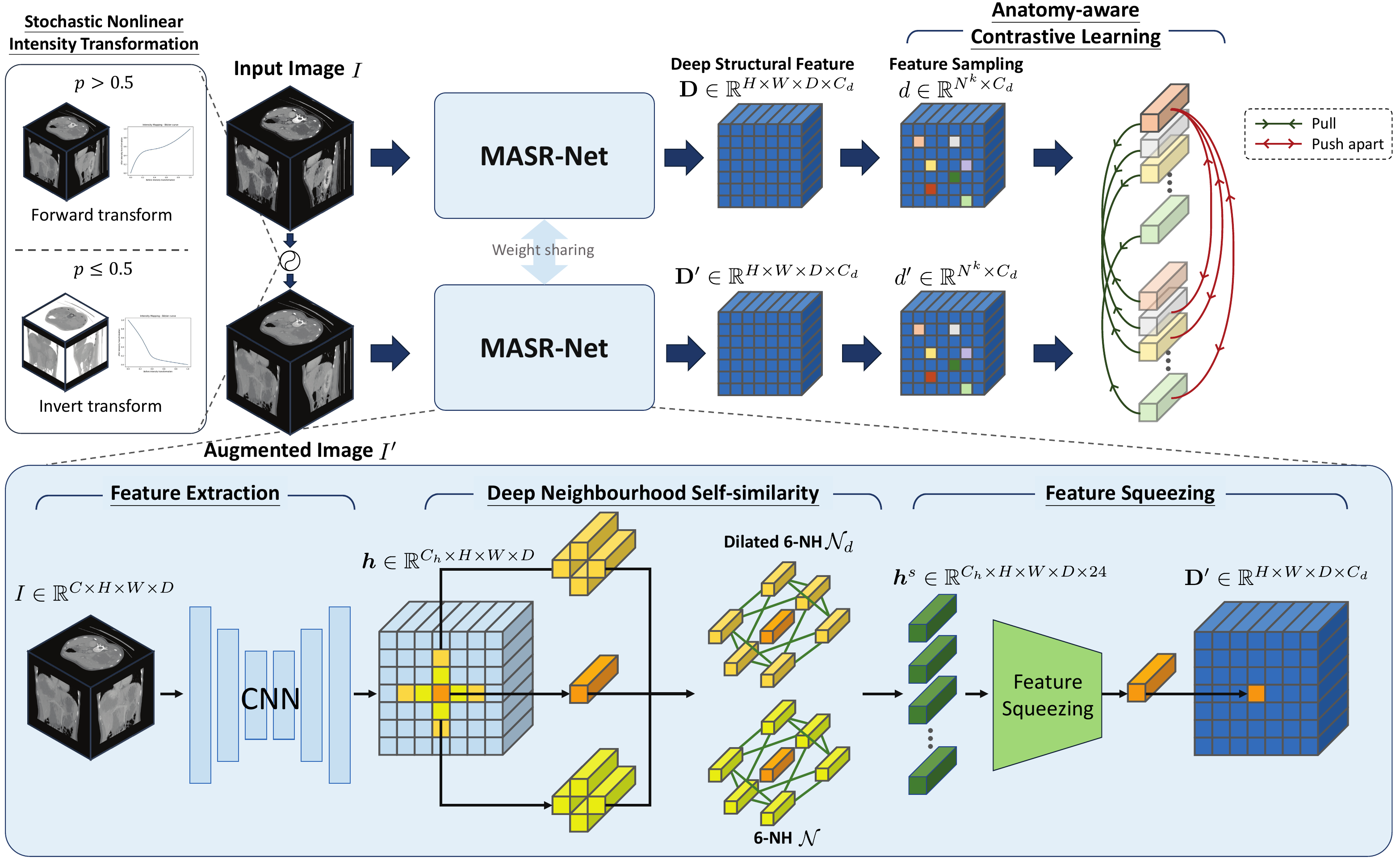}
	\end{center}
 \vspace{-1.em}
    \caption{Overview of the Modality-Agnostic Deep Structural Representation Network (MASR-Net) and anatomy-aware contrastive learning paradigm. For brevity, we visualize the 3D feature maps in a 2D aspect. Only negative pairs of the first feature vector are shown.}
    
	\label{fig:overview}
 \vspace{-12pt}
\end{figure*}


\subsection{Problem Setup and Overview}
Let $\F$, $\M$ be fixed and moving volumes defined over a $n$-D mutual spatial domain $\Omega \subseteq \mathbb{R}^n$. For simplicity, we further assume that $\F$ and $\M$ are three-dimensional, single-channel, and grayscale images, \ie, $n = 3$ and $\Omega \subseteq \mathbb{R}^3$. In this paper, we aim to extract DSIRs of $\F$ and $\M$, \ie,  $\mathbf{D}_\F$ and $\mathbf{D}_\M$, with high discriminability such that only the cosine similarity of two feature vectors in the identical anatomical location $x$, 
\ie, $\mathbf{D}_\F(x)$ and $\mathbf{D}_\M(x)$, or with similar structural information is maximised. To this end, we introduce a Modality-Agnostic deep Structural Representation Network (MASR-Net, Sec.~\ref{sec:modality_agnostic}) and anatomy-aware contrastive learning paradigm (Sec.~\ref{sec:contrastive}), followed by a multimodal similarity metric formulation with DNS (Sec.~\ref{sec:multimodal_sim}).

The overview of MASR-Net and anatomy-aware contrastive learning paradigm is illustrated in the lower and upper panels of Fig.~\ref{fig:overview}, respectively. Our network first computes the image feature with an encoder-decoder network, extracts the deep structural information from the feature map with the DNS extractor, and encodes them with the feature squeezing module. The proposed network is trained with non-linear intensity transformation, followed by anatomy-aware contrastive learning. With these components, the complex structural and anatomical location-aware information are well reflected in the resulting deep intermediate structural representation. 

\subsection{Modality-Agnostic Deep Structural Representation Network}\label{sec:modality_agnostic}

\paragraph{Feature extraction.}
We first leverage a feed-forward 3D convolutional neural network (CNN) to extract the image feature $\h$ for the input image $I$. The proposed CNN network is built with a 4-level encoder-decoder structure with skip connection \cite{ronneberger2015u}, which is composed of 3D convolution layers, LeakyReLU activations \cite{maas2013rectifier} and uses BlurPool \cite{zhang2019making} and trilinear interpolation for downsampling and upsampling, respectively, resulting in a maximum striding factor of 8. The network takes the input image $I\in\mathbb{R}^{C \times H \times W \times D}$ and outputs an image feature map $\h\in\mathbb{R}^{C_h \times H \times W \times D}$. The details of the network are shown in supplementary material.

\paragraph{Deep Neighbourhood Self-similarity (DNS).}
The vanilla local self-similarity descriptor \cite{shechtman2007matching} captures internal geometric layouts of local self-similarities within images by computing the pairwise distance between the patch center and its neighbourhood within a patch \emph{at the pixel-level}. Yet, this formulation may be sensitive to the noise presented in the image, especially for medical images. In contrast, our proposed DNS is computed \emph{at the feature level} and avoids using the patch center for the calculation, capable of capturing complex structural information of the image beyond the local neighbour context and less sensitive to the noise presented in medical images.

Formally, given the feature map $\h\in\mathbb{R}^{C_h\times H \times W \times D}$ of input image $I$, a patch centred at $x\in\Omega$ and a certain neighbourhood layout $\mathcal{N}$, the proposed deep neighbourhood self-similarity $\mathbf{S}$ is given by:
\begin{small} 
\begin{equation}
\begin{array}{l}
    \! \mathbf{S}(\h, x, y) \! = \! \exp \Big(\!- \! \sum_{y'\in\mathcal{N}(x)} \frac{(\h(y) - \h(y'))^2}{\sigma^2}\Big), y' \neq y,
\end{array}
\end{equation}
\end{small}
\noindent where $y, y' \in \mathcal{N}(x)$ defines the neighbour location of $x$. The denominator $\sigma^2$ is a noise estimator, defined as the mean of all patch distances, \ie, $\frac{1}{|\mathcal{N}(x)|}\sum_{y'\in\mathcal{N}(x)}(\h(y) - \h(y'))^2$, where $|\mathcal{N}(x)|$ represents the number of neighbour voxels in $\mathcal{N}(x)$. We follow \cite{heinrich2013towards} to further restrict the pairwise distance calculations within the six-neighbourhood (6-NH) with a Euclidean distance of $\sqrt{2}$ between them, reducing the computation complexity of DNS, \ie, reduce the number of unique pair-wise distances from $15$ to $12$. 

To further maximize the discriminability of the computed feature, we compute two sets of DNS from $\h$ using two different neighbourhood layouts, \ie, $\small \mathcal{N}$ and $\mathcal{N}_d$. We define $\mathcal{N}$ and $\mathcal{N}_d$ to be the direct 6-NH and dilated 6-NH layouts, respectively, as shown in Fig.~\ref{fig:overview}. The DNS of the two neighbourhood layouts is then concatenated to form a 5-D feature map $\h^s\in\mathbb{R}^{C_h \times H \times W \times D \times 24}$, containing the deep structural information of the feature map $\h$.  

\paragraph{Feature Squeezing.}
To compute a compact, intermediate DSIR from the 5-D DNS feature map $\h^s$, we encode the high-dimensional 5-D feature map $\h^s$ into a compact 4D DNS descriptor using a feature squeezing module. The feature squeezing module consists of a single-layer linear layer with $C_h$ perceptions, followed by a feed-forward convolution head. It first takes the 5-D DNS feature vector $\h^s$ as input and encodes it into a compact deep structural embedding $\h^c \in \mathbb{R}^{H \times W \times D \times 24}$ using linear projection. The feed-forward convolution head is composed of two 3D convolution layers (kernel size $=3^3$) with LeakyReLU activation in between the layers. It further encodes the compact deep structural embedding $\h^c$ to the DSIR $\mathbf{D} \in \mathbb{R}^{H \times W \times D \times C_d}$.

\subsection{Anatomy-aware Contrastive Learning}\label{sec:contrastive}
The deep intermediate structural representation $\mathbf{D}$ using DNS is a regional descriptor that is able to capture local geometric structures in the feature map while suppressing appearance variation inside it. However, most regions in the image may share similar local geometric structures or suffer from image noise, causing ambiguity in matching the true anatomical correspondence. To further enhance the discriminability of $\mathbf{D}$, we introduce an anatomy-aware contrastive learning paradigm that learns to maximize the similarity of the deep structural embeddings of the input image and augmented image at the identical anatomical locations while penalizing the similarity of that from different anatomical locations. Our anatomy-aware contrastive learning is depicted in the upper panel of Fig.~\ref{fig:overview}.

\paragraph{Stochastic non-linear intensity transformation.}
To simulate the heterogeneous intensity distribution of the anatomical structures or lesions across multi-modal scans, we perform stochastic non-linear intensity augmentation to the augmented images, as shown in Fig.~\ref{fig:overview}. Specifically, we employ the Bézier curve \cite{bezier_curve}, which is controlled by $n+1$ random control points $P_i \in \mathbb{R}^2$ uniformly sampled within $[0,1]$ and a monotonic nonlinear transformation, to augment the intensity of the augmented image. The monotonicity ensures a one-to-one correspondence between pixel values, thereby maintaining the geometric structures of the image. Formally, the Bézier curve \( B(t) \) is defined as:
\begin{small} 
\begin{equation}\label{eq:b_curve}
\begin{array}{l}
	B(t) = \sum_{i=0}^{n} P_i \cdot b_{i, n}(t),
\end{array}
\end{equation}
\end{small}
\noindent where \( P_i \) are the control points, \( b_{i,n}(t) \) are the Bernstein polynomials, \( n \) is the number of control points minus one, and \( t \) ranges from 0 to 1. Furthermore, to enforce the contrast invariant properties to the deep structural embeddings, we randomly invert the image intensity. The resulting nonlinear intensity transformation is then applied to the image intensity values as follows:
\begin{small} 
\begin{equation}\label{eq:nonlinear_mapping}
    I'(x) = 
\begin{cases}
    B(I(x)),& \text{if } p > \delta\\
    B(1-I(x)), & \text{if } p \le \delta
\end{cases}
\quad x \in \Omega,
\end{equation}
\end{small} 
\noindent where $p \in [0,1]$ is the randomly sampled probability and $\delta$ is the threshold for the intensity inversion. The stochastic non-linear intensity transformation diversifies the intensity distribution of the structural information in the training set while preserving the crucial structural information between input and augmented image, improving the robustness of the learned deep intermediate structural representation across different modalities. 

\paragraph{Contrastive loss.}
Let $\mathbf{D},\mathbf{D}'$ denote the DSIR of the input image $I$ and augmented image $I'$, respectively. For optimal contrastive learning, the goal is to maximize the similarity of $\mathbf{D}$ and $\mathbf{D}'$ at identical anatomical locations while penalizing the similarity of that from different anatomical locations. Since $I$ and $I'$ are originated from the same image, $\mathbf{D}(x), \mathbf{D}'(x)$ are the deep structural embeddings sampled at identical anatomical locations given any $x\in\Omega$ sampled in the foreground of $I$. 

For optimal contrastive learning, we randomly sample $N^k$ feature vectors from $\mathbf{D}$ and $\mathbf{D'}$, respectively. Let $d_i = ||\mathbf{D}(i)||_2$ and $d'_i = ||\mathbf{D}'(i)||_2$ be the unit vector of $D(i)$ sampled with spatial indice $i$, where $i = 1, \dots, N^k$. Indices in correspondence $d_i$ and $d_i^+$ are positive pairs. Similarly, $d_i$ and $d_j^-$, where $j = 1, \dots, N^k$ and $j \neq i$, are negative pairs. The contrastive loss is defined as:
\begin{small} 
\begin{equation}\label{eq:contrastive_learning}
\ell(d_i, d_i^+, d_j^-) \!= \!-\!\log{ \Bigg[
\frac{\exp (d_i \cdot d_i^+ / \tau)}{\exp (d_i \cdot d_i^+/\tau) + \sum_{j=1,j \neq i}^{N^k} \exp (d_i \cdot d_j^- /\tau)} \Bigg] }.
\end{equation}
\end{small}

\subsection{Multimodal Similarity Using DNS}\label{sec:multimodal_sim}
Without the loss of generality, given $\mathbf{D}_F$ and $\mathbf{D}_M$ are the DSIRs of $F$ and $M$ estimated by MASR-Net,  the multi-modal image registration problem with DNS can be formulated as:
\begin{small} 
\begin{equation}
\begin{array}{l}
  \!  \phi^{*} \! = \! \argmin_{\theta} \mathcal{L}_{sim}(\psi_\sigma(\mathbf{D}_F), \psi_\sigma(\mathbf{D}_M(\phi_\theta))) \! + \! \lambda\mathcal{L}_{reg}(\phi_\theta),
\end{array}
\end{equation}
\end{small}

\noindent where $\phi^*$ denotes the optimal displacement field, the $\psi_\sigma$ is a Gaussian smoothing function with variance $\sigma$, $\mathcal{L}_{sim}(\mathbf{D}_F, \mathbf{D}_M(\phi_\theta))$ denotes the negated cosine similarity function that quantifies the dissimilarity between the $\mathbf{D}_F$ and the $\mathbf{D}_M(\phi_\theta)$ spatially transformed by $\phi$, and $\mathcal{L}_{reg}(\phi)=||\nabla \phi||^2_2$ represents the smoothness regularization function that penalizes implausible solution. Our proposed DNS is differentiable and can be adapted to iterative instance optimization methods \cite{mok2023deformable,siebert2021fast} and learning-based registration frameworks \cite{BalakrishnanZSG19,mok2021conditional,mok2020fast,mok2022affine}. To exemplify the flexibility of DSIR, we parameterize the registration function with one learning-based method, LapIRN \cite{mok2021conditional}, and one 3-level non-rigid instance optimization method (IO) \cite{mok2022robust}, with implementation details shown in the supplementary material.

\begin{figure*}[!t]
\begin{minipage}[t]{1\textwidth}
\makeatletter\def\@captype{table}
\small
\centering
\resizebox{\textwidth}{!}{%
		\begin{tabular}{cccccccccccc}
			\toprule[1.5pt]
			\multirow{3}{*}{Method}  & \multirow{3}{*}{Metric}  & \multicolumn{5}{c}{Pre-contrast $\leftarrow$ Venous \& Arterial} & \multicolumn{5}{c}{Arterial $\leftarrow$ Venous \& Pre-contrast} \\
			\cmidrule(lr){3-7}\cmidrule(lr){8-12}
             & & \multicolumn{2}{c}{Tumour} & \multicolumn{2}{c}{Organ} & \multirow{2}{*}{$\%|J_\phi|<$0 $\downarrow$} & \multicolumn{2}{c}{Tumour} & \multicolumn{2}{c}{Organ} & \multirow{2}{*}{$\%|J_\phi|<$0 $\downarrow$}\\
            \cmidrule(lr){3-4}\cmidrule(lr){5-6}\cmidrule(lr){8-9}\cmidrule(lr){10-11}
			 & & \rule{1pt}{0ex} DSC $\uparrow$ & HD95 $\downarrow$ & \rule{1pt}{0ex} DSC $\uparrow$ & HD95 $\downarrow$ &  & \rule{1pt}{0ex} DSC $\uparrow$ & HD95 $\downarrow$ & \rule{1pt}{0ex} DSC $\uparrow$ & HD95 $\downarrow$ &  \\
			\midrule[1pt]
            Initial \hspace{0.1cm} & -- & 75.51 $\pm$ 20.64 & 4.12 $\pm$ 4.00 & 88.03 $\pm$ 8.77 & 3.80 $\pm$ 3.31 & -- & 77.84 $\pm$ 19.86 & 3.61 $\pm$ 3.19 & 90.15 $\pm$ 7.41 & 3.15 $\pm$ 2.68 & --  \\
            \midrule
            ANTs  \hspace{0.1cm} & MI & 76.52 $\pm$ 20.16 &  3.97 $\pm$ 3.15 & 87.40 $\pm$ 9.75 &  3.88 $\pm$ 2.91 &   0.00 $\pm$ 0.00&  80.18 $\pm$ 18.47 &  3.42 $\pm$ 2.73  & 90.07 $\pm$ 8.04 &  3.22  $\pm$ 2.80 & 0.00 $\pm$ 0.00  \\
            NiftyReg  \hspace{0.1cm} & MI & 77.10 $\pm$ 17.00 & 3.54 $\pm$ 2.54 &  90.87 $\pm$ 6.57 &  2.98 $\pm$ 2.10 & 0.35 $\pm$ 1.20 & 79.22 $\pm$ 16.54 & 3.18  $\pm$ 2.07 &  92.39 $\pm$ 5.61 &  2.53 $\pm$ 1.79 & 0.29 $\pm$ 1.18 \\
            DEEDs  \hspace{0.1cm} & MIND & 79.65 $\pm$ 15.30 & 3.14 $\pm$ 1.98 & 93.90 $\pm$ 3.47 & 2.09 $\pm$ 1.50 & 0.23 $\pm$ 0.72 & 80.57 $\pm$ 15.23 & 3.42 $\pm$ 2.06 & 94.80 $\pm$ 2.75 & \textbf{1.84} $\pm$ 1.04 & 0.12 $\pm$ 0.13 \\
            \midrule
            VM  \hspace{0.1cm} & NMI & 75.78 $\pm$ 17.35 & 3.62  $\pm$ 2.69 &  91.70 $\pm$ 4.43 &  2.81 $\pm$ 1.89 & 0.00 $\pm$ 0.00 & 76.62 $\pm$ 18.19 &  3.23 $\pm$ 2.30 &  92.69 $\pm$ 3.81  & 2.45 $\pm$ 1.63 & 0.00 $\pm$ 0.00 \\
            VM  \hspace{0.1cm} & MIND &  75.16 $\pm$ 17.07  & 3.94 $\pm$ 5.51 & 91.70 $\pm$ 4.92 &  2.77 $\pm$ 2.05 &  0.00 $\pm$ 0.00  & 75.18 $\pm$ 17.71 & 3.63 $\pm$ 5.11 & 92.15  $\pm$ 4.39 & 2.55 $\pm$ 1.65 & 0.00 $\pm$ 0.00 \\
            LapIRN  \hspace{0.1cm} & NMI & 78.80 $\pm$ 15.37 & 3.33 $\pm$ 2.80 &  93.53 $\pm$ 3.75 &  2.35 $\pm$ 1.52 &  0.01 $\pm$ 0.02  &  80.17 $\pm$ 15.38 &  2.87 $\pm$ 2.00  &  94.48 $\pm$ 3.72  &  1.97 $\pm$ 1.53  & 0.02 $\pm$ 0.16  \\
            LapIRN  \hspace{0.1cm} & MIND & 77.50 $\pm$ 16.71 & 3.96 $\pm$ 3.28 & 93.32 $\pm$ 4.41 & 2.46 $\pm$ 1.81 & 0.00 $\pm$ 0.01  & 79.49 $\pm$ 15.00 & 3.19 $\pm$ 2.44 & 94.29 $\pm$ 4.19 & 2.07 $\pm$ 1.77 & 0.01 $\pm$ 0.04 \\
            LapIRN (ours) \hspace{0.1cm} & DNS & 79.72 $\pm$ 14.44 &  3.06 $\pm$ 2.23 &  94.07 $\pm$ 3.36 & \textbf{2.08} $\pm$ 1.45 & 0.00 $\pm$ 0.01 &  80.66 $\pm$ 14.56 & \textbf{2.71} $\pm$ 1.85 & 94.73 $\pm$ 3.09 & 1.84 $\pm$ 1.25 & 0.02 $\pm$ 0.18 \\
            \midrule
            IO \hspace{0.1cm} & MIND & 76.27 $\pm$ 16.44 &  3.66 $\pm$ 2.74 & 92.54 $\pm$ 3.41 & 2.63 $\pm$ 1.32 & 0.08 $\pm$ 0.37 & 76.91 $\pm$ 16.14 & 3.54 $\pm$ 2.23 & 92.74 $\pm$ 3.45 & 2.70 $\pm$ 1.31 & 0.12 $\pm$ 0.51 \\
            IO (ours) \hspace{0.1cm} & DNS &  \textbf{80.43} $\pm$ 13.72  &  \textbf{2.94} $\pm$ 2.23  & \textbf{94.26} $\pm$ 3.32 & 2.10 $\pm$ 1.50  &  0.03 $\pm$ 0.20  &  \textbf{81.07} $\pm$ 13.83  &  2.74 $\pm$ 1.80  &  \textbf{94.89} $\pm$ 2.92  &  1.85 $\pm$ 1.22  & 0.06 $\pm$ 0.35  \\
			\bottomrule[1.5pt]
    \end{tabular} }
\end{minipage}
\qquad

\begin{minipage}[t]{1\textwidth}
\makeatletter\def\@captype{table}
\small
\centering
\resizebox{\textwidth}{!}{%
		\begin{tabular}{cccccccccccc}
			\toprule[1.5pt]
			\multirow{3}{*}{Method} & \multirow{3}{*}{Metric} & \multicolumn{5}{c}{Venous $\leftarrow$ Arterial \& Pre-contrast} & \multicolumn{5}{c}{\multirow{2}{*}{Average Score across Three Tasks}} \\
			\cmidrule(lr){3-7}
            & & \multicolumn{2}{c}{Tumour} & \multicolumn{2}{c}{Organ}  & \multirow{2}{*}{$\%|J_\phi|<$0 $\downarrow$} &  &  &  & & \\
            \cmidrule(lr){3-4}\cmidrule(lr){5-6} \cmidrule(lr){8-12}
			& & \rule{1pt}{0ex} DSC $\uparrow$ & HD95 $\downarrow$ & \rule{1pt}{0ex} DSC $\uparrow$ & HD95 $\downarrow$ &   & \rule{1pt}{0ex} DSC $\uparrow$ & HD95 $\downarrow$ & $\%|J_\phi|<$0 $\downarrow$ & $\textnormal{T}_{Test}$ & \# Param \\
			\midrule[1pt]
             Initial \hspace{0.1cm} & -- & 78.10 $\pm$ 19.85 & 3.59 $\pm$ 3.07  & 88.96 $\pm$ 7.49 & 3.50 $\pm$ 2.41 & --  & 83.10 $\pm$ 12.60 & 3.63 $\pm$ 2.97 & --  & -- & -- \\
			\midrule
            ANTs  \hspace{0.1cm} & MI & 81.14 $\pm$ 17.14  & 3.37 $\pm$ 2.69 &  89.20 $\pm$ 8.28 &  3.52 $\pm$ 2.46  &  0.00 $\pm$ 0.00 &  84.09  $\pm$ 12.84 &  3.56 $\pm$ 2.60  &  0.00 $\pm$ 0.00  & 250.05 $\pm$ 367.42* &  --    \\
            NiftyReg  \hspace{0.1cm} & MI & 78.53 $\pm$ 17.05 & 3.32 $\pm$ 2.28 &  91.19 $\pm$ 6.80 &  2.90 $\pm$ 2.06 & 0.32 $\pm$ 1.19 & 84.88 $\pm$ 9.92 & 3.07 $\pm$ 1.98  &  0.32 $\pm$ 1.19 & 79.17 $\pm$ 30.77*  &  --  \\
            DEEDs  \hspace{0.1cm} & MIND & 81.28 $\pm$ 14.65 & 3.32 $\pm$ 1.87  & \textbf{94.58} $\pm$ 2.44  & \textbf{2.11} $\pm$ 1.14 & 0.11 $\pm$ 0.14 & 87.46 $\pm$ 8.97 & 2.65 $\pm$ 1.72 & 0.15 $\pm$ 0.43 & 47.92 $\pm$ 12.77*  & --  \\
            \midrule
            VM  \hspace{0.1cm} & NMI &  77.20 $\pm$ 17.33 & 3.28  $\pm$ 2.09  & 92.65 $\pm$ 3.31  &  2.60 $\pm$ 1.43 & 0.00 $\pm$ 0.00  & 84.44 $\pm$ 9.47 & 3.00  $\pm$ 1.83 & 0.00 $\pm$ 0.00 & 0.14 $\pm$ 0.01   & 1.14M  \\
            VM  \hspace{0.1cm} & MIND & 75.94 $\pm$ 17.18 & 3.34 $\pm$ 1.54 &  92.25 $\pm$ 4.07 &  2.63 $\pm$ 1.54 &  0.00 $\pm$ 0.00 &  83.73 $\pm$ 9.33 &  3.14 $\pm$ 2.57 & 0.00 $\pm$ 0.00 & 0.16 $\pm$ 0.01   & 1.14M  \\
            LapIRN  \hspace{0.1cm} & NMI & 81.37 $\pm$ 14.05 &  2.86 $\pm$ 2.12 & 94.22 $\pm$ 3.46 & 2.24 $\pm$ 1.57 & 0.00 $\pm$ 0.00 & 87.10 $\pm$ 8.89 & 2.61 $\pm$ 1.91 & 0.00 $\pm$ 0.01 &  0.19 $\pm$ 0.01 &  1.59M   \\
            LapIRN  \hspace{0.1cm}& MIND & 81.03 $\pm$ 14.56 & 3.09 $\pm$ 2.61  & 93.99 $\pm$ 3.98 & 2.33 $\pm$ 1.82 &  0.00 $\pm$ 0.00 &  86.61 $\pm$ 9.81 & 2.85 $\pm$ 2.29 &  0.00 $\pm$ 0.01 &  0.18 $\pm$ 0.01 &  1.59M   \\
            LapIRN (ours)  \hspace{0.1cm} & DNS &  \textbf{81.62} $\pm$ 13.42 &  2.81 $\pm$ 1.99 & 94.37 $\pm$ 3.03 & 2.08 $\pm$ 1.41 & 0.00 $\pm$ 0.00 &  87.53 $\pm$ 8.65 & 2.43 $\pm$ 1.70 & 0.01 $\pm$ 0.01 &  0.18 $\pm$ 0.01  &  1.59M    \\
            \midrule
            IO \hspace{0.1cm} & MIND &  77.44 $\pm$ 16.26 & 3.48 $\pm$ 2.37 & 93.02 $\pm$ 3.30 & 2.56 $\pm$ 1.42 & 0.01 $\pm$ 0.02 & 84.82 $\pm$ 9.83 & 3.10 $\pm$ 1.90 & 0.07 $\pm$ 0.30 & 4.75 $\pm$ 0.44 &  -  \\
            IO (ours)  \hspace{0.1cm} & DNS &  81.51 $\pm$ 13.67  &  \textbf{2.79} $\pm$ 1.94 &  94.53 $\pm$ 3.01  &  2.11 $\pm$ 1.48 &  0.00 $\pm$ 0.01  &  \textbf{87.78} $\pm$ 8.41  &  \textbf{2.42} $\pm$ 1.70 &  0.03 $\pm$ 0.19  &  5.05 $\pm$ 0.47  &  -   \\
			\bottomrule[1.5pt]
    \end{tabular} }
\caption{Quantitative results on liver multi-phase CT registration task. $X \leftarrow Y$ represents the experiment of registering the CT scan in the $Y$ phase to the CT scan in the $X$ phase of the same patient. $\uparrow$: higher is better, and $\downarrow$: lower is better. Initial: initial results without registration. The registration runtime highlighted with an asterisk is reported in CPU time (in seconds). }
\label{tab:main_result}
\end{minipage}
\vspace{-12pt}
\end{figure*}

\section{Experiments}

\subsection{Data and Pre-processing}
\paragraph{Liver Multiphase CT. }~We collected a 3D multi-phase contrast-enhanced liver CT dataset from hospitals in China to evaluate the \emph{inter-patient registration}. This dataset consists of 1966 patients with tumors, and each patient contains scans in three phases, \ie, pre-contrast, arterial and venous phases, acquired from the CT scanner at different time points after the contrast agent injection. There are large non-linear deformations and heterogeneous image intensities across multi-phase CT due to gravity, muscle contractions, and contrast agent injection. 
We divided the dataset into 1886/10/70 cases for training, validation, and test sets. The liver and tumour masks of the validation and test sets are manually annotated and checked by a senior radiologist.

\paragraph{Abdomen MR-CT. } We evaluated our method on the Abdomen MR-CT task of the Learn2Reg challenge 2021~\citep{hering2022learn2reg}. The dataset comprises 8 sets of paired MR and CT scans and 90 unpaired CT/MR scans, both depicting the abdominal region of a single patient and presenting notable deformations. Since there are only limited paired MR and CT scans, we only evaluate the iterative optimization-based registration methods in this task. We use all 8 paired MR and CT scans as the test set. Each scan has four manually labeled anatomical structures: liver, spleen, right kidney, and left kidney. For MASR-Net, we split the 90 unpaired CT/MR scans into 85 and 5 scans for training and validation sets.

\paragraph{Brain MR T1w-T2w. } We performed atlas-based registration on BraTS18~\footnote{https://www.med.upenn.edu/sbia/brats2018.html} and ISeg19~\footnote{https://iseg2019.web.unc.edu/} datasets~\cite{6975210,SunGWLZLWMYFZPS21}.
The training set consists of 135 cases, each with T1-weighted contrast (T1w) and T2-weighted contrast (T2w) images, and was resampled to 1mm$^3$ isotropic resolution. 10 cases contain full delineation of anatomical structures, including grey matter, white matter and cerebrospinal fluid (CSF) for both T1w and T2w modalities. We randomly select 1 case from the test set as the atlas and register the remaining 9 cases to the atlas.
%
%
Similar to Abdomen MR-CT, we only evaluate iterative optimization-based registration methods on this task due to the limited training data.

\begin{figure*}[t]
	\centering
    \includegraphics[width=1.0\textwidth]{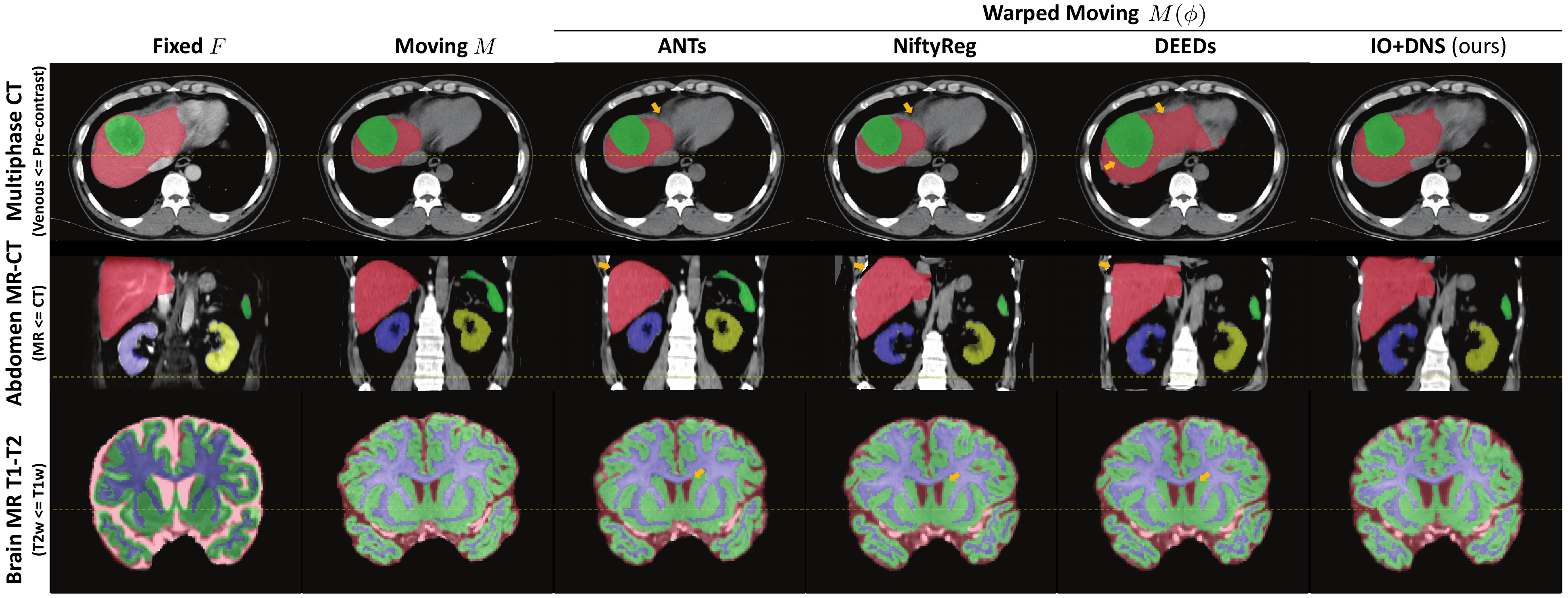}
    \caption{ Example slices of resulting warped images using different registration methods on Liver multiphase CT, Abdomen MR-CT, and Brain MR T1w-T2w registration tasks. The warped anatomical segmentations are overlayed on the resulting images.  Major registration artefacts are highlighted with yellow arrows. Multiphase CT: tumour (green) and liver (red). Abdomen MR-CT: spleen (green), liver (red), right kidney (yellow), and left kidney (blue). MR T1w-T2w: grey matter (green), white matter (blue) and cerebrospinal fluid (red).
    }
	\label{fig:qualitative}
 \vspace{-10pt}
\end{figure*}

\begin{table}[t]
	\begin{center}
	\centering
     \resizebox{0.48\textwidth}{!}{%
		\begin{tabular}{ccccccc}
			\toprule[1.5pt]
			\multirow{2}{*}{Method}   & \multirow{2}{*}{Metric} & \multicolumn{2}{c}{Abdomen MR $\leftarrow$ CT} & \multicolumn{3}{c}{Brain MR T1w $\leftrightarrow$ T2w}\\
            \cmidrule(lr){3-4}\cmidrule(lr){5-7}
			 & & \rule{1pt}{0ex} DSC $\uparrow$ & $\textnormal{T}_{Test}$ & \rule{1pt}{0ex} DSC$_{\text{T1} \leftarrow \text{T2}}$ $\uparrow$  & \rule{1pt}{0ex} DSC$_{\text{T2} \leftarrow \text{T1}}$ $\uparrow$ &  $\textnormal{T}_{Test}$  \\
			\midrule[1pt]
			Initial \hspace{0.1cm} &  -- & 37.32 $\pm$ 17.23  &  -- & 53.90 $\pm$ 0.70   &  53.90 $\pm$ 0.70  &  -- \\
                \midrule
                DEEDs \hspace{0.1cm} & MIND & 83.53 $\pm$ 8.58   & 165.21 $\pm$ 25.14* & 61.47 $\pm$ 0.96 & 60.70 $\pm$ 0.83 &  14.45 $\pm$ 0.54 \\
                ANTs \hspace{0.1cm} &  MI & 39.43 $\pm$ 17.67  & 38.54 $\pm$ 3.51* & 61.00 $\pm$ 1.00   &  52.80 $\pm$ 1.10  &  18.46 $\pm$ 1.33\\
                NiftyReg \hspace{0.1cm} & MI & 48.60 $\pm$ 31.60 & 37.88 $\pm$ 10.95* & 63.90 $\pm$ 1.10  &  61.90 $\pm$ 0.70  & 34.75 $\pm$ 3.04\\
                \midrule
                IO (ours) \hspace{0.1cm} & DNS & 85.61 $\pm$ 5.95  &  4.52 $\pm$ 0.50 & 62.66 $\pm$ 0.73 &  61.91 $\pm$ 0.66  &  5.84 $\pm$ 0.54\\
			\bottomrule[1.5pt]
		\end{tabular}
	}
	\caption{Quantitative results on Abdomen MR-CT and brain MR T1w-T2w registration tasks. $\uparrow$: higher is better, and $\downarrow$: lower is better. Initial: initial results in native space without registration.}
 \vspace{-1.5em}
	\label{tab:main_result_add}
\end{center}
\end{table}

\subsection{Implementation}
All DLIR methods, \ie, DNS, Voxelmorph, and LapIRN, are developed and trained using Pytorch. All the methods are trained or executed on a standalone workstation equipped with 8 Nvidia V100 GPUs and an Intel Xeon Platinum 8163 CPU. We adopt the Adam optimizer \cite{kingma2014adam} with a fixed learning rate of $1e^{-4}$ and batch size set to 1 for all learning-based approaches. For the training of MASR-Net, we set the number of sampling feature vectors $N^k$, temperature $\tau$, the number of random control points $n$, threshold $\delta$ and $C_d$ to 8196, 0.07, 3, 0.5 and 24, respectively. For the IO method, we adopt the 3-level multiresolution optimization strategy and Adam optimizer with a fixed learning rate of $1e^{-3}$ for each level. 

\subsection{Evaluation metrics and Baselines}
To quantify the registration performance, we register each pair, propagate the anatomical segmentation map using the resulting transformation, and measure the region of interest (ROI) segmentation overlap using the Dice similarity coefficient (DSC). We also measure the 95\% percentile of the Hausdorff distance (HD95) to represent the registration accuracy to the boundary of the ROI structures. As a negative determinant of the Jacobian at a voxel indicates local folding~\cite{Ashburner07}, we further compute the percentage of foldings ($\%|J_\phi|<$0) within the deformation field to evaluate the plausibility of the deformation fields.

We compare our method with three state-of-the-art conventional registration methods (ANTs~\cite{AvantsTSCKG11}, NiftyReg~\cite{SunNK14} and DEEDs~\cite{HeinrichJBS12}) and two learning-based approaches (Voxelmorph~\cite{BalakrishnanZSG19} and LapIRN~\cite{MokC20}). Specifically, we use the ANTs registration implementation in the publicly available ANTs software package~\cite{avants2009advanced}. Both ANTs and NiftyReg methods use a four-level multi-resolution strategy with adaptive gradient descent optimization and MI as the similarity measure. 
For Voxelmorph and LapIRN, we use official implementations and the best hyperparameter reported in their paper.
By default, all learning-based methods are trained with two similarity metrics (NMI and MIND).

\begin{figure*}[t]
	\begin{center}
            \includegraphics[width=.85\linewidth]{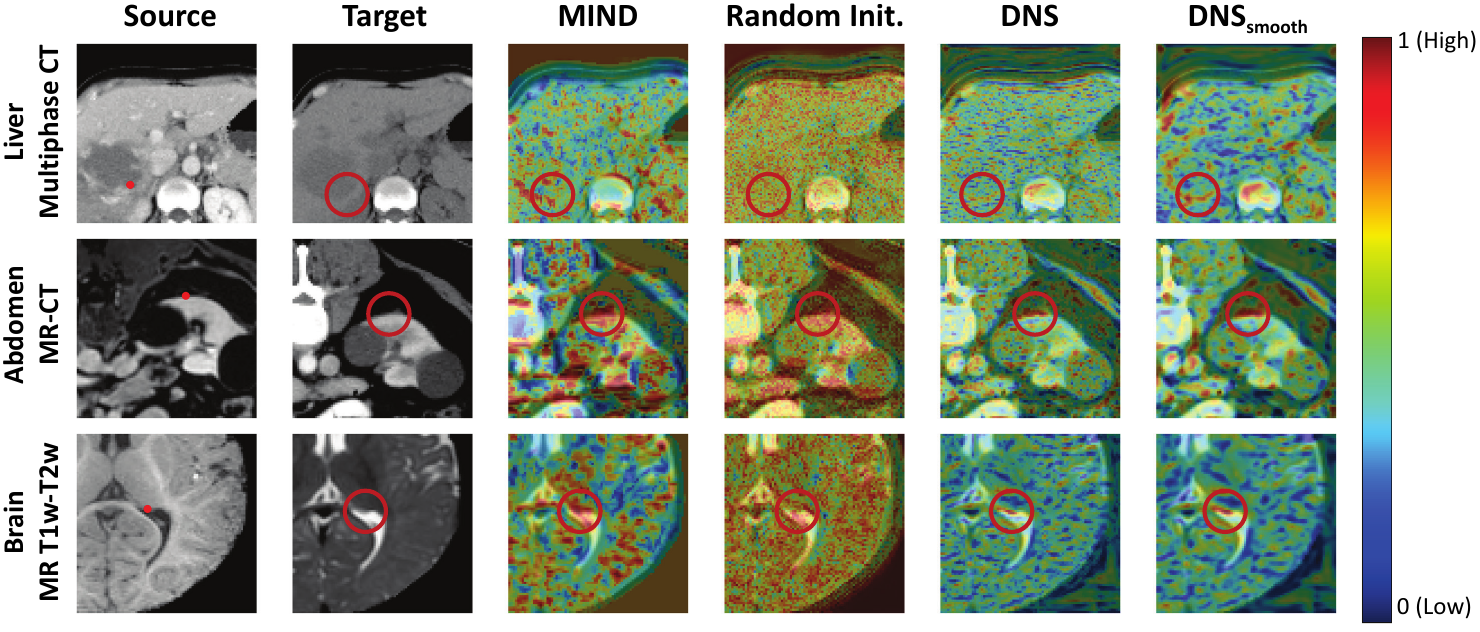}
	\end{center}
   \vspace{-1.5em}
    \caption{Visualization of the feature similarity maps across different modalities and anatomies. Random Init.: DNS without training. DNS$_{\text{smooth}}$: DNS with Gaussian smoothing. Each heatmap shows the similarity of the marked point (red pot) on the source image to every point in the target image. The feature extracted with our method (DNS and DNS$_{\text{smooth}}$) shows high discriminability at the boundary of liver tumour ($1^{\text{st}}$ row), kidney ($2^{\text{nd}}$ row) and anatomical structure of brain MR with large intensity variation ($3^{\text{rd}}$ row).}
	\label{fig:feature_similarity}
 \vspace{-12pt}
\end{figure*}

\subsection{Results}
\paragraph{Intra-subject Liver multiphase CT registration.}
Table~\ref{tab:main_result} shows the results of the Liver multiphase CT registration. The initial DSC of the registration of venous and arterial CT to pre-contrast (Pre-contrast $\leftarrow$ Venous \& Arterial) is the lowest among all directions, implying that there is a relatively large misalignment between pre-contrast and enhanced CT scans. Overall, our proposed method (IO+DNS and LapIRN+DNS) achieves the best registration accuracy in terms of DSC and HD95 over all three registration directions, suggesting that our method is robust and accurate in multiphase CT registration, especially in the alignment of tumor region, as shown in the first row of Fig.~\ref{fig:qualitative}. 

\noindent\emph{Flexibility and versatility.} Compared to the results of LapIRN with different metrics, LapIRN with our DNS achieves the best overall registration performance in tumor and organ alignment, showing +2.22 and +0.75 DSC gains over the one with MIND in tumor and organ alignment, respectively. More importantly, when comparing the results of IO with DNS and MIND, there are significant improvements in registration accuracy of both tumor and organ (+4.16 and + 1.72 gains in DSC). The result proves that our method is more robust and flexible with diverse registration frameworks than MIND. When using DNS with a simple iterative gradient optimization, the DSIR of DNS is more discriminative than MIND and can lead to less ambiguous matching during the iterative optimization. Besides, although NMI shows comparable DSC in multiphase CT registration, the qualitative results in Fig.~\ref{fig:qualitative} demonstrate that methods using MI/NMI fail to recover the large non-linear misalignment in MR-CT registration.

%
%
%
%
%

\begin{table}[t]
	\begin{center}
		\scalebox{1.0}{
			\resizebox{0.47\textwidth}{!}{
					\begin{tabular}{l|ll}
						\toprule
					Methods     & DSC $\uparrow$ & HD95 $\downarrow$ \\
                    \hline
     			Initial & 81.77 $\pm$ 16.96 & 3.96 $\pm$ 3.66  \\ 
                    \hline
					Backbone network feature (Random initialization)  & 75.38 $\pm$ 21.03 & 5.90 $\pm$ 5.46  \\ 
					\ \ \ \ \small{+Deep Neighbourhood Self-similarity}    & 86.50 $\pm$ 13.43 \color{ForestGreen}\small \textbf{(+11.12)}   & 2.79 $\pm$ 2.30  \color{ForestGreen}\small \textbf{(-3.11)}  \\
					\ \ \ \ \small{+Nonlinear Intensity \& Contrastive Learning}    & 87.13 $\pm$ 12.06 \color{ForestGreen}\small \textbf{(+0.63)}   & 2.57 $\pm$ 1.95 \color{ForestGreen}\small \textbf{(-0.22)}  \\
                    \ \ \ \ \small{+Gaussian Smoothing}    & 87.35 $\pm$ 12.14 \color{ForestGreen}\small \textbf{(+0.22)}   & 2.52 $\pm$ 1.95 \color{ForestGreen}\small \textbf{(-0.05)}    \\

	               \bottomrule
						\end{tabular}
					}}
	\end{center}
 \vspace{-1.5em}
	\caption{Influence of the deep neighbourhood self-similarity and contrastive learning to the registration model.}
  \vspace{-1.5em}
	\label{tab:ablation}
\end{table}

\vspace{-12pt}
\paragraph{Intra-subject abdomen MR-CT registration.}
Table \ref{tab:main_result_add} depicts the result of abdomen MR-CT registration. 

\noindent\emph{Robust to large deformation.} The initial DSC of this task is relatively low (37.32), indicating there are possible large spatial misalignments between image pairs. All conventional methods (ANTs and NiftyReg) with MI fail spectacularly in this task, suggesting that MI cannot register multimodal images with large deformation. In contrast to MI, structural image representation-based methods (DEEDs and DNS) achieve decent results in this task (83.53 and 85.61 DSC, respectively). It is worth noting that our DNS with a solely simple gradient descent-based optimization strategy (IO) outperforms DEEDs, which employ a more sophisticated convex optimization.

\vspace{-12pt}
\paragraph{Inter-subject brain MR T1w-T2w registration.}
Table \ref{tab:main_result_add} shows the results of brain MR T1w-T2w registration. 

\noindent\emph{Contrast invariance.} Brain MR T1w and T2w scans show a huge heterogeneous intensity distribution between anatomical structures across T1w and T2w modalities. Our method (IO+DNS) achieves on-par registration accuracy with conventional methods (DEEDs, ANTs and NiftReg), boosting the initial DSC from 53.90, 53.90 to 62.66 and 61.91, respectively, indicating that DNS is robust to contrast change in multimodal images. The feature similarity of DNS, as shown in Fig.~\ref{fig:feature_similarity}, further suggests that DNS has high discriminability and is robust to images with heterogeneous intensity distribution.

\vspace{-10pt}
\subsubsection{Ablation Studies}


\emph{Discriminability of the structural image representations.}
As demonstrated in Fig.~\ref{fig:feature_similarity}, each heatmap shows the similarity of the marked point on the source image to every point in the target image on three multi-modal registration tasks. Our method produces less ambiguous heatmaps than MIND on tumor boundaries ($1^{\text{st}}$ row) and anatomical structures ($2^{\text{nd}}$ and $3^{\text{rd}}$ rows), indicating our method is capable of capturing expressive and discriminative DSIR.

\noindent\emph{Effect of Deep Neighbourhood Self-similarity and contrastive learning.}
Tab.~\ref{tab:ablation} demonstrates performance metrics for different registration models on the multiphase CT registration task (Pre-contrast $\leftarrow$ Venous \& Arterial). Although a recent study \cite{frankle2020training} reported that random features from CNN could be expressive, the DSIR of random initialize CNN has lower DSC than that of initial, suggesting that random initialization alone is insufficient for multi-modal registration. Moreover, there is a significant performance gain when the DNS module is added to the randomly initialized CNN (+11.12 DSC), suggesting the effectiveness of DNS in extracting deep structural information. Comparing the feature similarity analysis of random initialization (MASR-Net without training) and DNS in Fig.~\ref{fig:feature_similarity}, one can observe that the proposed anatomy-aware contrastive learning effectively improves the discriminability of the DSIR of DNS, resulting in high feature similarity centralized in the true anatomical correspondence and relatively low feature similarity at the other anatomical structures.

\section{Conclusion}
This paper proposes a deep structural image representation learning method dedicated to multi-modality medical image registration. 
Our method leverages deep neighbourhood self-similarity to learn highly discriminative, contrast invariance structural representations for multimodal images. Anatomy-aware contrastive learning is introduced to further enhance the expressiveness and discriminability of the learned structural representation, reducing the ambiguity in matching the true anatomical correspondence between multimodal images. Comprehensive experiments demonstrate that our method achieves state-of-the-art registration accuracy compared with conventional local structural representations without the need for a dedicated optimizer.
{
    \small
    \bibliographystyle{ieeenat_fullname}
    \bibliography{main}
}

\clearpage
\setcounter{page}{1}
\maketitlesupplementary


\renewcommand*{\thesection}{\Alph{section}}
\setcounter{section}{0}

\section{Network Architecture}
\begin{figure}[h]
	\begin{center}
        \includegraphics[width=0.8\linewidth]{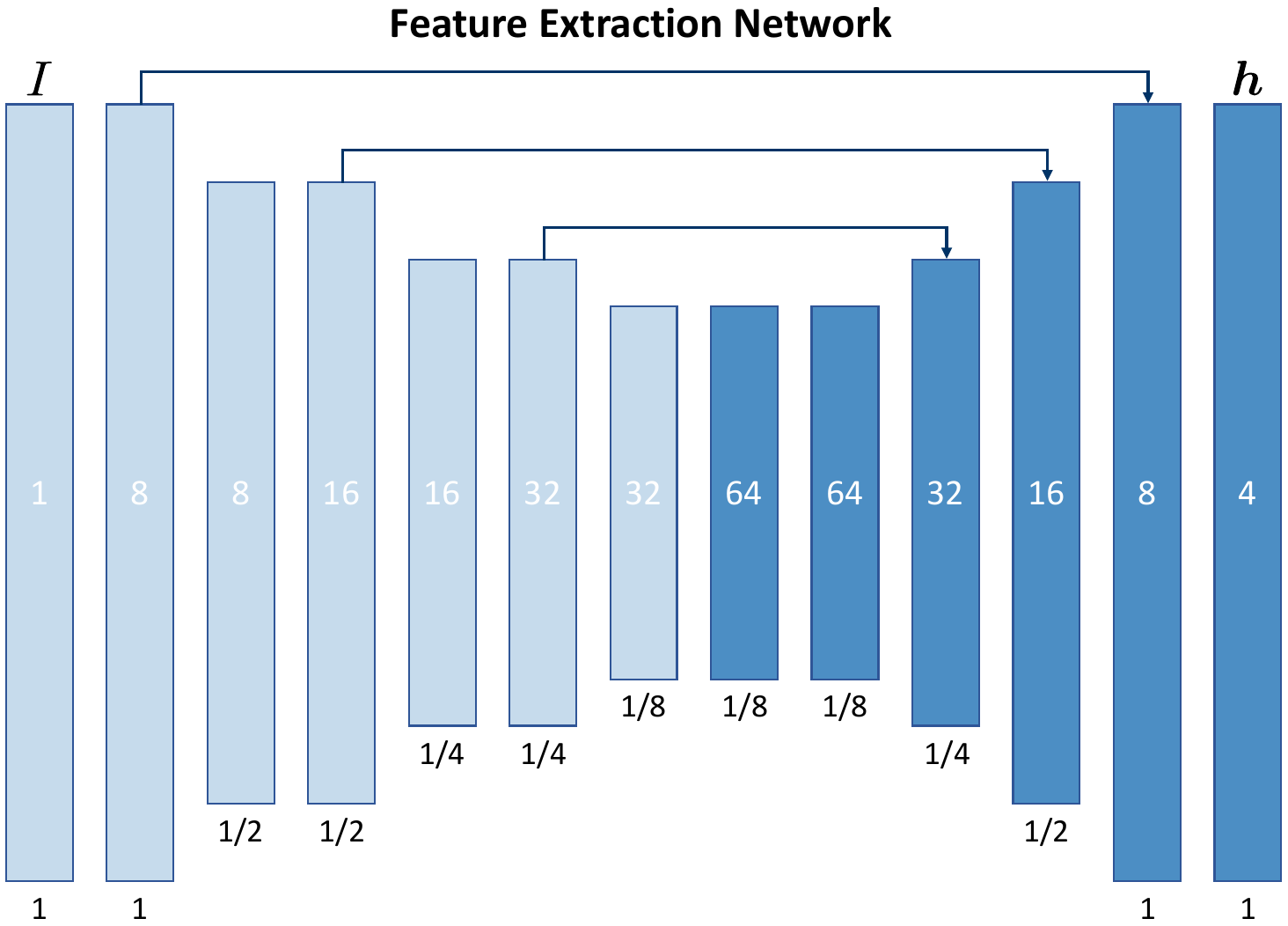}
	\end{center}
  \vspace{-1.4em}
	\caption{Network Architecture of the feature extraction network. Each rectangle denotes a 3D volume. The number of channels is shown inside the rectangle, and the spatial resolution with respect to the input volume is depicted underneath. The proposed feature extraction network uses BlurPool \cite{zhang2019making} and trilinear interpolation for downsampling and upsampling, respectively.}
	\label{fig:network_ffn}
\end{figure}

\begin{figure}[h]
	\begin{center}
        \includegraphics[width=1.0\linewidth]{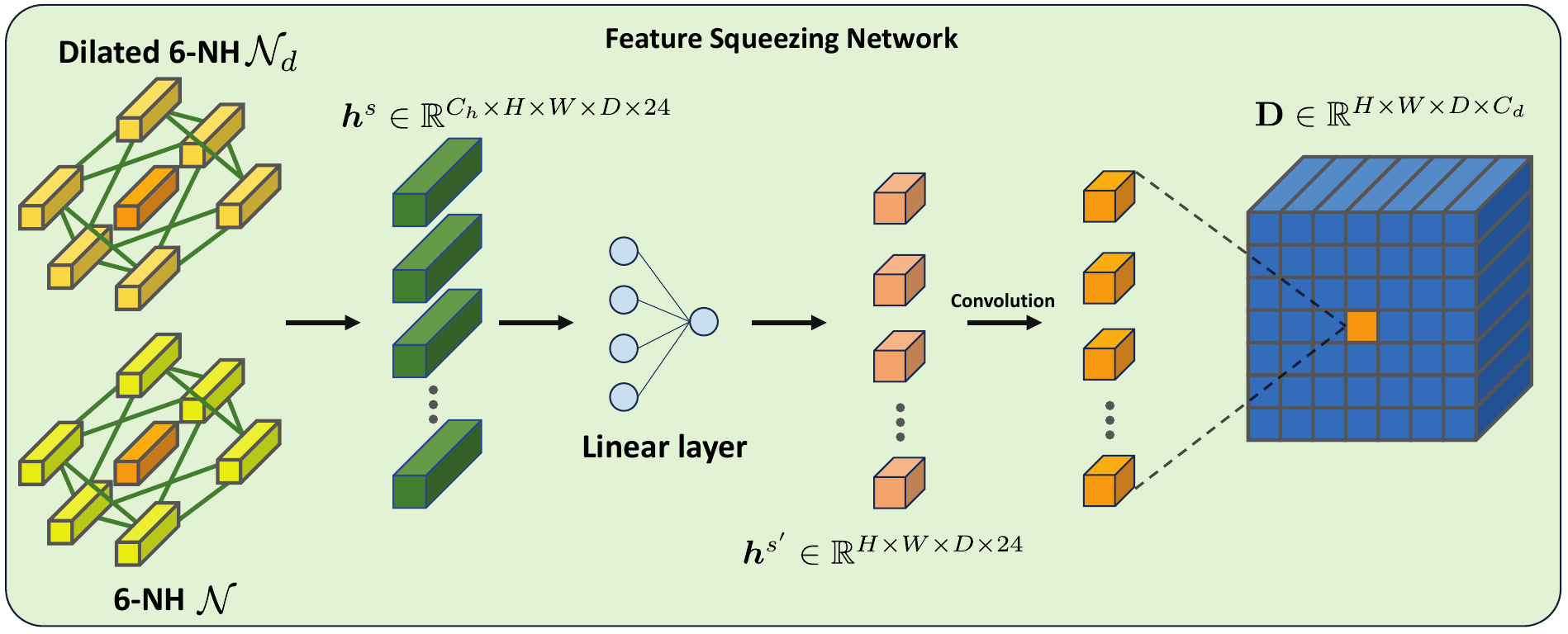}
	\end{center}
 \vspace{-1.4em}
	\caption{Network Architecture of the feature squeezing network. The feature squeezing network encodes the feature of deep neighbourhood similarity $\h^{s}$ into a compact deep structural image representation $\mathbf{D}$. The feed-forward convolution head is composed of two 3D convolution layers (kernel size $=3^3$) with LeakyReLU activation in between the layers.}
	\label{fig:network_fsn}
\end{figure}

Figures \ref{fig:network_ffn} and \ref{fig:network_fsn} depict the network architecture of the feature extraction and feature squeezing, respectively.

\section{Unsupervised Deformable Image Registration with Deep Structural Representations}
\subsection{Instance-specific Optimization (IO)}
\begin{figure*}[t]
	\begin{center}
        \includegraphics[width=0.85\linewidth]{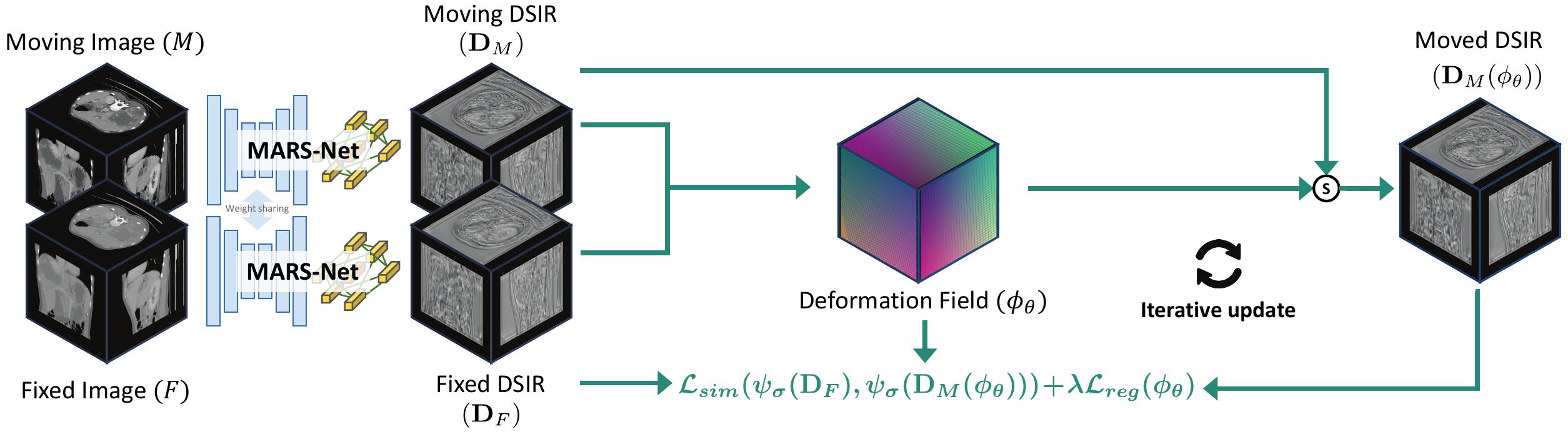}
	\end{center}
  \vspace{-1.2em}
	\caption{Schematic representation of the Instance-specific Optimization (IO) registration scheme with the proposed deep structural image representations. For brevity, only a 1-level IO registration scheme is shown.}
	\label{fig:training_IO}
\end{figure*}

Figure \ref{fig:training_IO} depicts the proposed instance-specific optimization with deep structural image representations. Note that the deep structural image representations are only required to be computed once for each input image throughout the registration. The pyramid of the deep structural image representations is computed by downsampling the deep structural image representations at full resolution with trilinear interpolation. Table \ref{tab:IO_parameters} shows the hyperparameters of the IO registration framework used for all registration tasks. 

\begin{table}[h]
\centering
\resizebox{0.47\textwidth}{!}{
\begin{tabular}{c|c}
 \toprule
\textbf{Parameters}  & \textbf{IO Model} \\ 
         \midrule 
Number of pyramid levels $N_{\text{level}}$      &     3       \\
Max. image dimension $N_{\text{max}}$        & $(H, W, D)$     \\
Min. image dimension $N_{\text{min}}$       & $(\frac{H}{2}, \frac{W}{2}, \frac{D}{2})$       \\
Learning rate per level  &  [1$e$-2, 5$e$-3, 3$e$-3]      \\
Max. iteration per level $N_{\text{iter}}$   &    [100, 80, 50]      \\
Max. number of grid points & $H \times W \times D$ \\
Min. number of grid points  & $ \frac{H}{2} \times \frac{W}{2} \times \frac{D}{2}$ \\
Weight of smoothness $\lambda_{reg}$  &  [0.6, 0.5, 0.4] \\
\bottomrule
\end{tabular}
}
\caption{Parameters used in IO registration framework. $H$, $W$ and $D$ refer to the height, width and depth of the input volume. }\label{tab:IO_parameters}
\end{table}

\subsection{Learning-based Image Registration}
\begin{figure*}[t]
	\begin{center}
        \includegraphics[width=.85\linewidth]{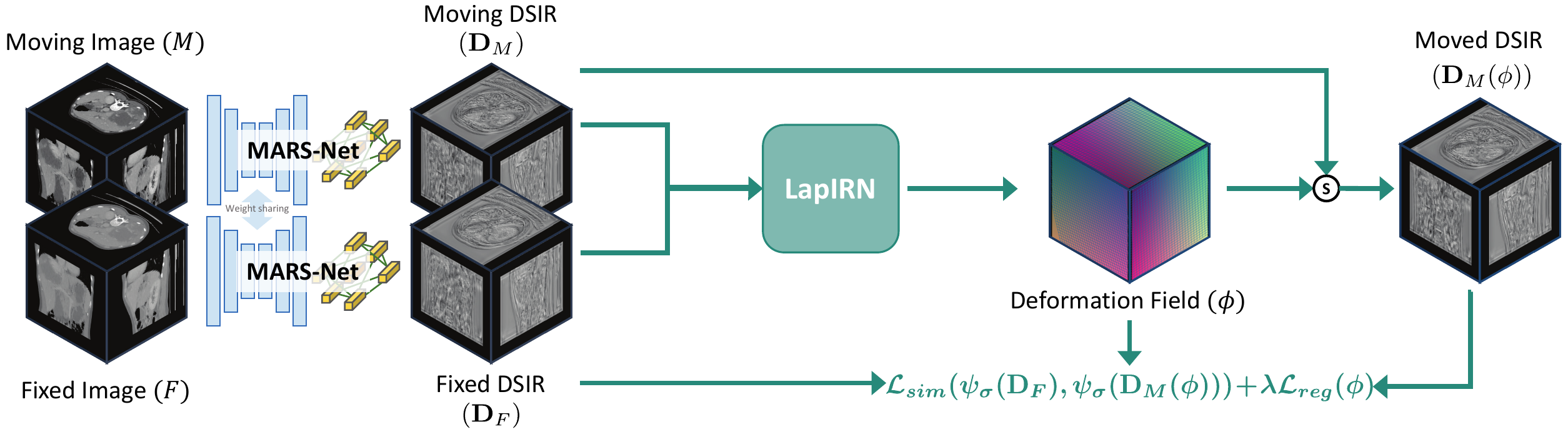}
	\end{center}
        \vspace{-1.2em}
	\caption{Schematic representation of the learning-based registration (LapIRN) scheme with the proposed deep structural representations.}
	\label{fig:training_LapIRN}
\end{figure*}

Figure \ref{fig:training_LapIRN} illustrates the example of a learning-based image registration method with deep structural image representations. We instantiate the learning-based image registration method with LapIRN \cite{mok2021conditional} throughout this paper. We highlight that our proposed method is flexible and can be adapted to other existing learning-based and iterative optimization-based methods with minimal effort.

\begin{table}[t]
	\begin{center}
		\scalebox{1.0}{
			\resizebox{0.47\textwidth}{!}{
					\begin{tabular}{l|ll}
						\toprule
					Methods     & DSC $\uparrow$ & HD95 $\downarrow$ \\
                    \hline
     			Initial & 37.32 $\pm$ 17.23 & 19.14 $\pm$ 9.54  \\ 
                    \hline
					MASR-Net w/o DNS  & 63.78 $\pm$ 17.44 & 11.15 $\pm$ 4.48 \\ 
					\ \ \ \ \small{DNS with single layout ($\mathcal{N}$)}    & 84.09 $\pm$ 6.89    & 5.54 $\pm$ 2.91   \\
					\ \ \ \ \small{DNS with dilated layout ($\mathcal{N}_d$)}    & 84.20 $\pm$ 8.24   & 5.48 $\pm$ 3.43   \\
					\ \ \ \ \small{DNS with duo layout ($\mathcal{N} \ \& \ \mathcal{N}_d$)}    & 85.79 $\pm$ 5.95   & 4.52 $\pm$ 2.39  \\

	               \bottomrule
						\end{tabular}
					}}
	\end{center}
 \vspace{-1.2em}
	\caption{Influence of the neighbourhood layout of deep neighbourhood self-similarity on the Abdomen CT-MR registration task using instance optimization as registration framework.}
	\label{tab:ablation_DNS}
\end{table}

\begin{figure*}[htp]
	\begin{center}
        \includegraphics[width=0.85\linewidth]{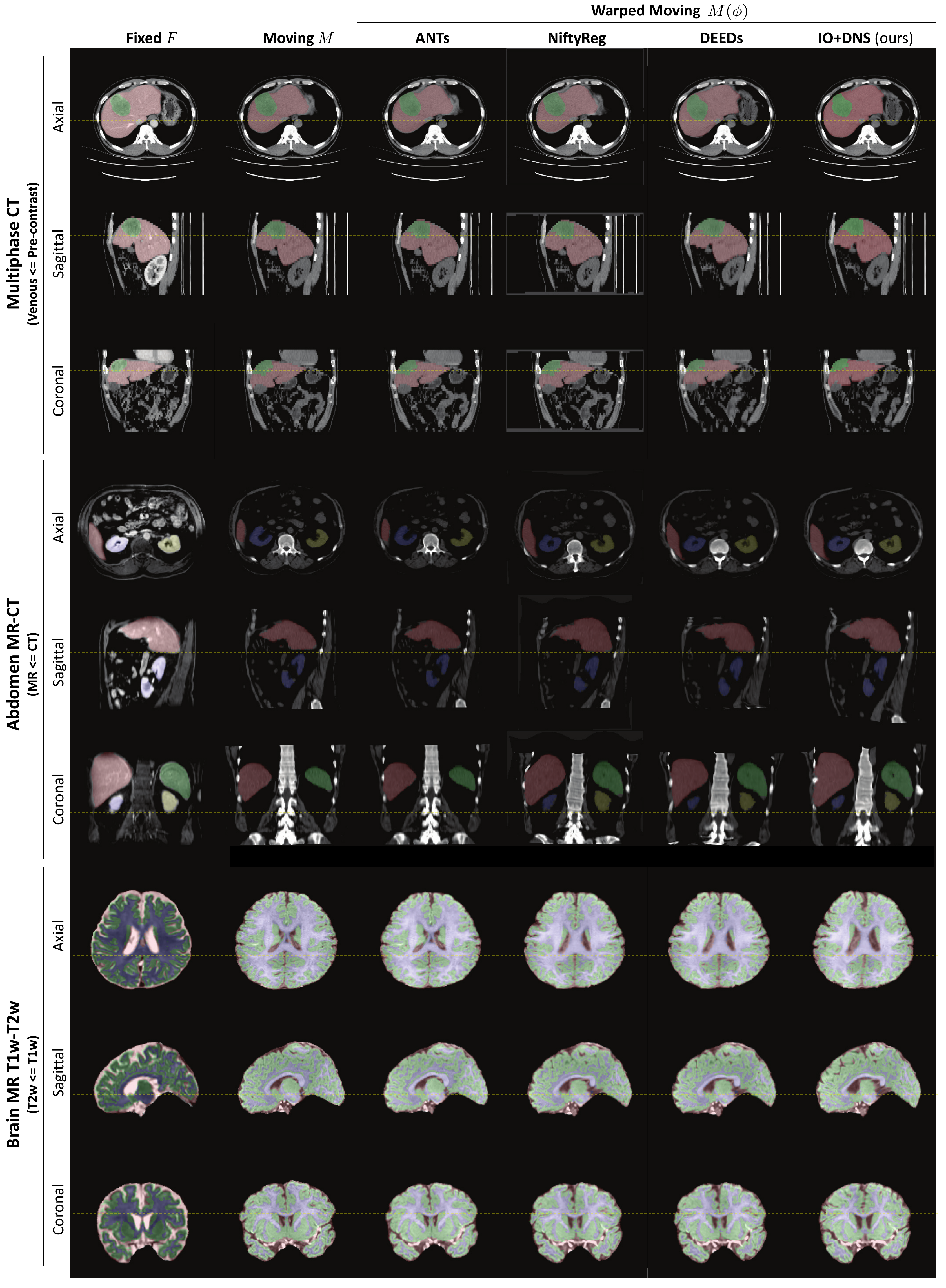}
	\end{center}
	\caption{Example axial, sagittal and coronal slices of resulting warped images using different registration methods on Liver multiphase CT, Abdomen MR-CT, and Brain MR T1w-T2w registration tasks. The warped anatomical segmentations are overlayed on the resulting images. Dotted yellow lines are added to highlight major misalignments of the registration. Multiphase CT: tumour (green) and liver (red). Abdomen MR-CT: spleen (green), liver (red), right kidney (yellow), and left kidney (blue). MR T1w-T2w: grey matter (green), white matter (blue) and cerebrospinal fluid (red).}
	\label{fig:additional_qual_registration}
\end{figure*}

\begin{figure*}[htp]
	\begin{center}
        \includegraphics[width=1.0\linewidth]{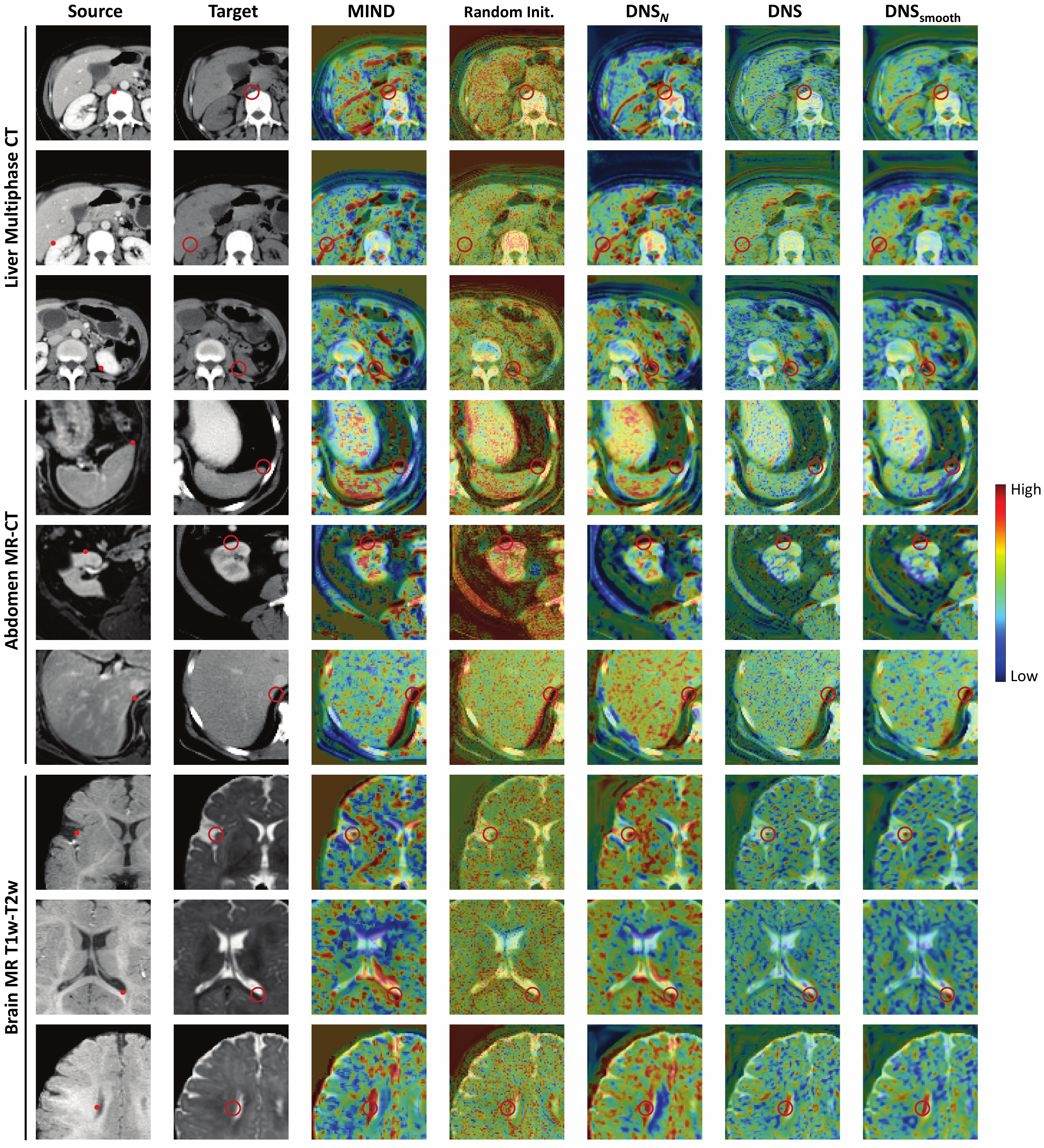}
	\end{center}
    \vspace{-1.5em}
	\caption{Visualization of feature similarity between the marked feature vector (red dot) of the source image and all feature vectors of the target image using different methods. MIND: Modality Independent Neighbourhood Descriptor. Random Init.: DNS without training. DNS$_\mathcal{N}$: DNS with single neighbourhood layout $\mathcal{N}$. DNS$_{\text{smooth}}$: DNS with Gaussian smoothing. The true anatomical correspondence of the marked point is highlighted with the red circle.}
	\label{fig:additional_feature_sim}
 \vspace{-1.5em}
\end{figure*}

\section{Details of Baselines and Data pre-processing}
\noindent The command and parameters we used for ANTs~\cite{AvantsTSCKG11}: \\
\begin{footnotesize}
\texttt{ANTS -d 3 -t Syn[0.25]
\\ -m MI[<Fixed>,<Moving>,1,32] 
\\ --number-of-affine-iterations 0x0 
\\-i 200x100x50 
\\ -r Gauss[9,0.2] 
\\-o <OutFileSpec>}
\end{footnotesize}\\

\noindent The command and parameters we used for NiftyReg~\cite{SunNK14}: \\
\begin{footnotesize}
\texttt{reg\_f3d -ref [<Fixed>] -flo [<Moving>]  
\\-cpp <OutFileSpec> -res <OutWarpedFile>} 
\end{footnotesize}\\

\noindent The command and parameters we used for DEEDs~\cite{HeinrichJBS12}: \\
\begin{footnotesize}
\texttt{ deedsBCV -F [<Fixed>] -M [<Moving>] 
\\ -O <OutFileSpec> -S <InSegmentationFile>
}
\end{footnotesize}\\

\paragraph{Liver Multiphase CT. }
Multiphase CT scans are cropped to the same field of view and resampled to an isotropic in-plane resolution of the $192\times192$ axial slices, with a slice thickness and increment from 2.0mm to 3.1mm. We further apply CT windowing to each image, clipping the Hounsfield (HU) value from -900 to 1024.

\paragraph{Abdomen MR-CT. }
For the CT scans in the abdomen MR-CT task, images are pre-aligned to the mutual space with size $192\times160\times192$ ($2\text{mm}^3$ isotropic resolution). We further apply CT windowing to each CT scan, clipping the Hounsfield (HU) value from -200 to 1024.


\section{Additional Experimental Results}
Figures \ref{fig:additional_qual_registration} and \ref{fig:additional_feature_sim} depict the registration results and feature similarity of the multiphase CT, abdomen MR-CT and brain MR T1w-T2w tasks.

\subsection{Non-linear Intensity Transformation}
\begin{figure*}[ht]
	\begin{center}
        \includegraphics[width=1.0\linewidth]{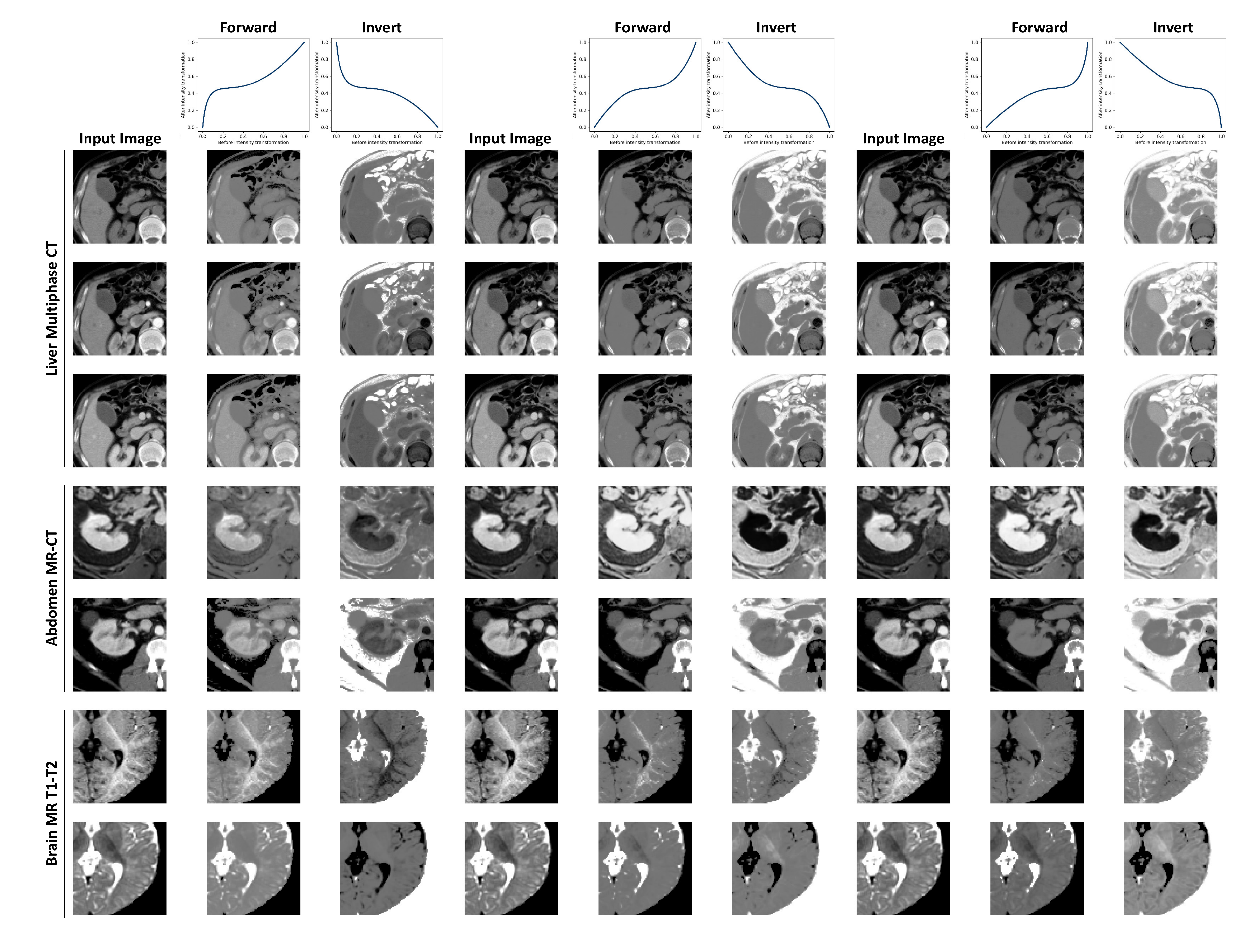}
	\end{center}
  \vspace{-1.5em}
	\caption{Example of the proposed stochastic non-linear intensity transformation with multiphase CT, abdomen MR-CT and brain MR T1w-T2w scans.}
	\label{fig:stochastic_intensity_transform}
\end{figure*}

Example images with the proposed stochastic non-linear intensity transformation are shown in Figure \ref{fig:stochastic_intensity_transform}.

\subsection{Liver Multiphase CT Registration}
\begin{figure*}[ht]
	\begin{center}
        \includegraphics[width=0.9\linewidth]{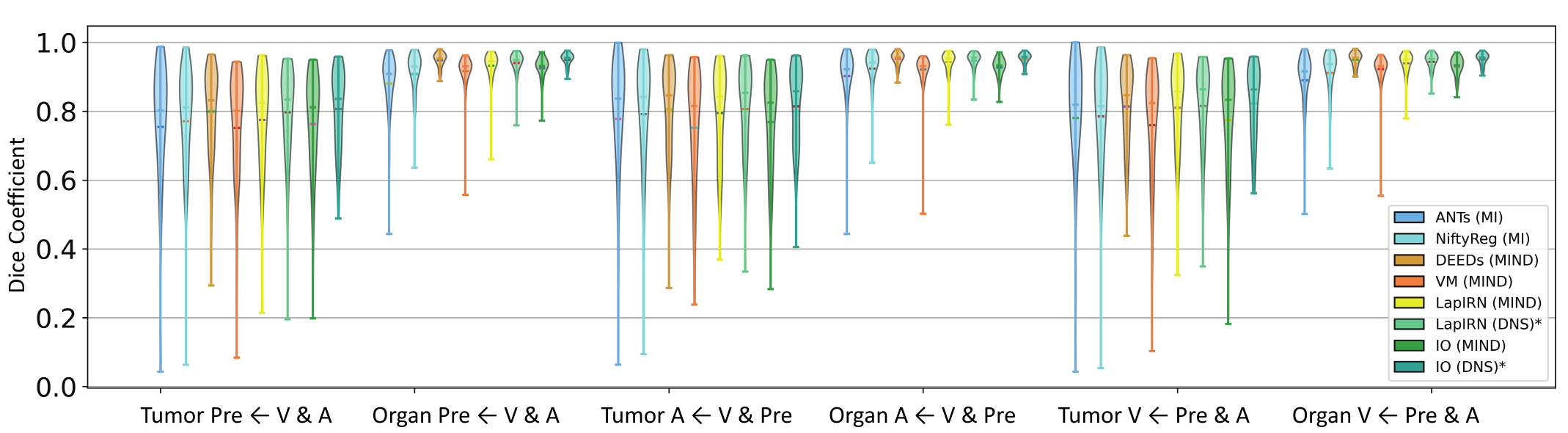}
	\end{center}
   \vspace{-1.5em}
	\caption{Boxplots of Liver Multiphase CT registration task. "Tumor Pre $\leftarrow$ V \& A" denotes the dice score of the tumour of registering venous and arterial to pre-contrast CT, while "Organ Pre $\leftarrow$ V \& A" represents the dice score of the liver of registering venous and arterial to pre-contrast CT.}
	\label{fig:additional_boxplot}
    \vspace{-1.em}
\end{figure*}

Figure \ref{fig:additional_boxplot} presents a boxplot of the result of the Liver Multiphase CT registration task in six registration directions. Our method utilizes a simple IO registration framework and achieves superior results over conventional methods with MI and MIND as similarity measures. Note that DEEDs use a convex optimization instead of gradient descent-based optimization, which is less sensitive to the sub-optimal solutions.

\subsection{Ablation study on DNS}
Table \ref{tab:ablation_DNS} shows the results of our proposed method with different combinations of neighbourhood layouts. MASR-Net with duo neighbourhood layout ($\mathcal{N} \ \& \ \mathcal{N}_d$) achieves the best overall result in the Abdomen CT-MR registration task, gaining +1.7\% and +1.59\% DSC improvement over MASR-Net with single neighbourhood layout $\mathcal{N}$ and $\mathcal{N}_d$, respectively.

\section{Tests of Statistical Significance}
Among the three directions in Liver multiphase CT registration, (P$\leftarrow$A \& V) registration is the most challenging task due to the large time interval of the image acquisition between fixed and moving images. IO$_\text{DNS}$ and LapIRN$_\text{DNS}$ achieve significant improvement ($p < 0.05$) in Dice (tumour and liver) over both learning-based (VM$_\text{MI}$, VM$_\text{MIND}$, LapIRN$_\text{MIND}$) and conventional methods (ANTs, NiftyReg). Results of the two-tailed t-test for MR-CT and T1w-T2w registration are provided in Table \ref{tab:p_value}. Notably, our method (IO-DNS and LapIRN-DNS) achieved statistically significant differences in Dice with IO-MIND, LapIRN-MIND and LapIRN-NMI, demonstrating the flexibility and effectiveness of DNS. 

\begin{table}[h]
	\begin{center}
	\centering
     \resizebox{0.48\textwidth}{!}{%
		\begin{tabular}{ccccc}
			\toprule[1.5pt]
            Method   & Metric & Abdomen$_{\text{MR} \leftarrow \text{CT}}$ & Brain MR$_{\text{T1w} \leftarrow \text{T2w}}$ & Brain MR$_{\text{T2w} \leftarrow \text{T1w}}$\\
			\midrule[1pt]
            DEEDs & MIND & $\bm{p<0.05}$ & $\bm{p<0.001}$ & $\bm{p<0.05}$ \\
            ANTs & MI & $\bm{p<0.001}$ & $\bm{p<0.05}$ & $\bm{p<0.001}$ \\
            NiftyReg & MI & $\bm{p<0.001}$ & $\bm{p<0.05}$ & $p>0.10$ \\
			\bottomrule[1.5pt]
		\end{tabular}
	}
	\caption{Significance level for baseline methods in comparison to IO$_\text{DNS}$ (ours). Results with \emph{significant differences} to IO$_\text{DNS}$ (ours) are highlighted in \textbf{bold}.}
	\label{tab:p_value}
\end{center}
\end{table}

\begin{figure*}[t]
	\begin{center}
        \includegraphics[width=0.75\linewidth]{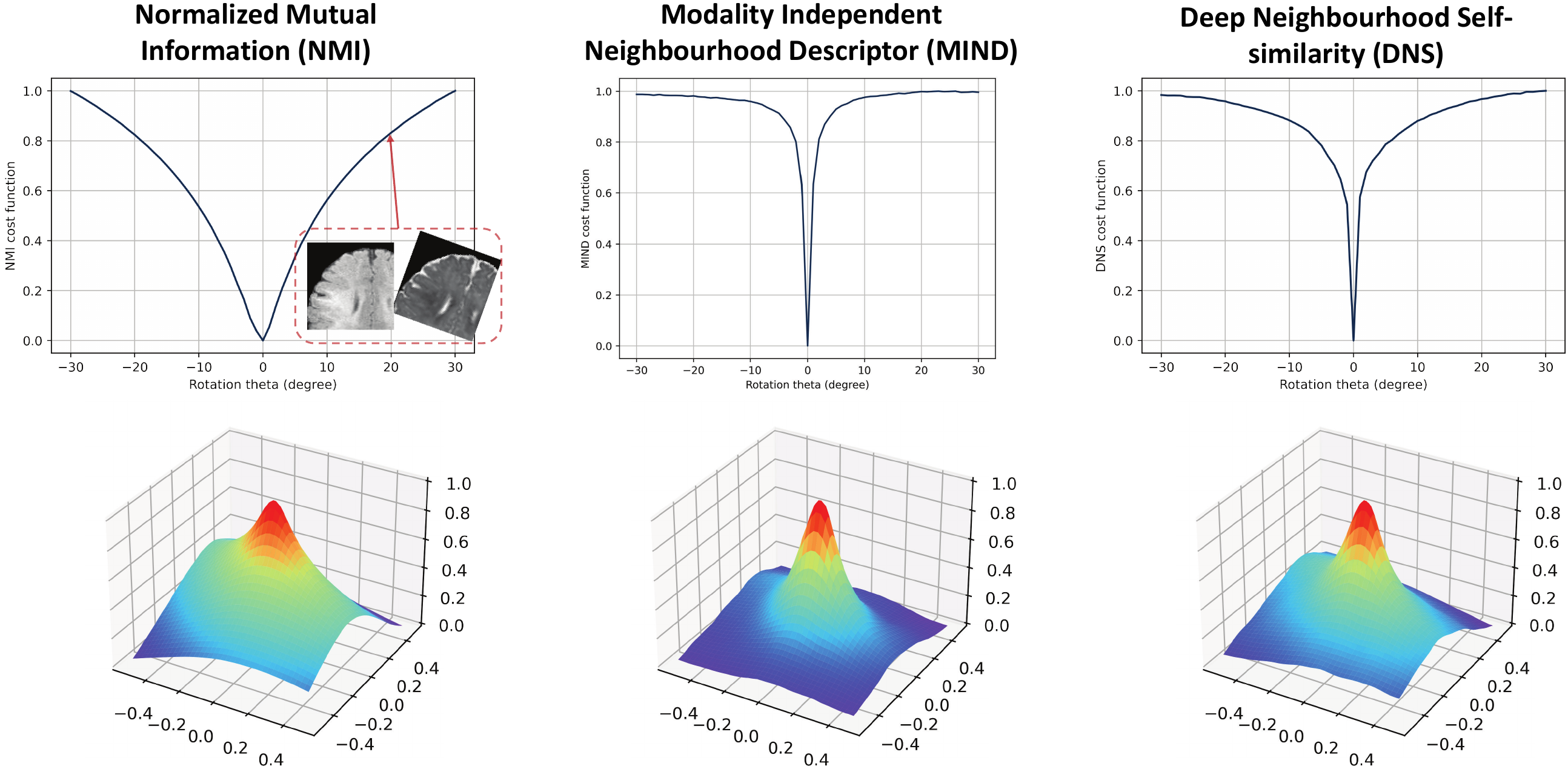}
	\end{center}

    \caption{Loss landscapes of NMI, MIND and DNS (ours) on T1w-T2w MRI image with varying rotations along two axes.}\label{fig:rotation}

	\label{fig: compare}
\end{figure*}

\section{Loss Landscapes}
The loss landscapes on co-registered MR T1w and T2w with varying degrees of rotation along two axes are shown in Fig. \ref{fig:rotation}. Comparing to NMI and MIND, DNS demonstrates a smooth transition of cost function for high rotation degrees while persevering sharp minimum loss the images are perfectly aligned, indicating that DNS possesses high discriminability and sensitivity to linear shifts in multi-modal images. 

\end{document}